\documentclass[review,authoryear]{elsarticle}

\usepackage{framed,multirow}
\usepackage{amsmath,amssymb,amsfonts}
\usepackage{latexsym}
\usepackage{algorithmic}
\usepackage{textcomp}
\usepackage{booktabs}          
\usepackage{nicefrac}       
\usepackage{graphicx,wrapfig}
\graphicspath{{Figs/}}
\usepackage{subfig}
\usepackage{tikz}
\usepackage{comment}
\usepackage{multirow}

\DeclareMathOperator*{\argmin}{arg\,min}

\newcommand*\diff{\mathop{}\!\mathrm{d}}
\usepackage{xcolor}
\usepackage{url}
\usepackage{hyperref}

\begin{document}
\begin{frontmatter}

\title{NPB-REC: A Non-parametric Bayesian Deep-learning Approach for Undersampled MRI Reconstruction with Uncertainty Estimation}%

\author[1]{Samah Khawaled\corref{cor1}}
\cortext[cor1]{Corresponding author}
\ead{ssamahkh@campus.technion.ac.il}
\author[2]{Moti Freiman \corref{cor2}}
\ead{moti.freiman@technion.ac.il}

\address[1]{the Interdisciplinary program in Applied Mathematics, Faculty of Mathematics, Technion – Israel Institute of Technology}
\address[2]{the Faculty of Biomedical Engineering, Technion – Israel Institute of Technology }

\begin{abstract}
 The ability to reconstruct high-quality images from undersampled  MRI data is vital in improving MRI temporal resolution and reducing acquisition times. Deep learning methods have been proposed for this task, but the lack of verified methods to quantify the uncertainty in the reconstructed images hampered clinical applicability.
 We introduce ``NPB-REC'', a non-parametric fully Bayesian framework, for MRI reconstruction from undersampled data with uncertainty estimation. We use Stochastic Gradient Langevin Dynamics during training to characterize the posterior distribution of the network parameters. This enables us to both improve the quality of the reconstructed images and quantify the uncertainty in the reconstructed images. We demonstrate the efficacy of our approach on a multi-coil MRI dataset from the fastMRI challenge and compare it to the baseline End-to-End Variational Network (E2E-VarNet).
 Our approach outperforms the baseline in terms of reconstruction accuracy by means of PSNR and SSIM ($34.55$, $0.908$ vs.  $33.08$, $0.897$, $p<0.01$, acceleration rate $R=8$) and provides uncertainty measures that correlate better with the reconstruction error (Pearson correlation, $R=0.94$ vs. $R=0.91$). Additionally, our approach exhibits better generalization capabilities against anatomical distribution shifts (PSNR and SSIM of $32.38$, $0.849$ vs. $31.63$, $0.836$, $p<0.01$, training on brain data, inference on knee data, acceleration rate $R=8$). 
 NPB-REC has the potential to facilitate the safe utilization of deep learning-based methods for MRI reconstruction from undersampled data. Code and trained models are available at \url{https://github.com/samahkh/NPB-REC}.

\end{abstract}


\end{frontmatter}

\section{Introduction}
\label{sec:introduction}

Magnetic resonance imaging (MRI) is a non-invasive imaging modality that provides multi-planar images through its sensitivity to the inherent magnetic properties of human tissue \cite{morris2018magnetic}. Despite its excellent sensitivity to soft tissue contrast, non-invasiveness, and lack of harmful ionizing radiation, long acquisition times remain a significant challenge in achieving high spatial and temporal resolutions, reducing motion artifacts, improving the patient experience, and lowering costs \cite{fastmri}.

Undersampling the ``k-space'' (Fourier domain) in MRI acquisition can reduce acquisition time and enable advanced MRI applications, but it also results in aliasing artifacts in the reconstructed images. Early approaches relied on Parallel Imaging (PI) to recover high-quality images by utilizing multiple receiver coils simultaneously to acquire the data and then combine the acquired data linearly to construct a high-quality image \cite{griswold2002generalized, pruessmann1999sense, sodickson1997simultaneous,sarasaen2021fine}. Non-linear compressed sensing (CS) approaches leverage sparsity prior coupled with non-linear optimization to reconstruct the MRI image from the undersampled data \cite{candes2006compressive, lustig2007sparse}. However, conventional CS approaches only provide a point estimate of the reconstructed image. Recently, Jalal et al. proposed Compressed Sensing with Generative Models (CSGM) to characterize the posterior distribution via Langevin dynamics \cite{jalal2021robust}. However, this approach requires iterative solutions during reconstruction rather than a single forward pass of the deep neural network (DNN).

In recent years, Deep Neural Networks (DNN)-based models have surpassed classical reconstruction approaches by reconstructing high-quality MRI images from highly under-sampled data (i.e., 25\% or less) \cite{shaul2020subsampled,edupuganti2020uncertainty,eo2018kiki,akccakaya2019scan,tezcan2018mr,morris2018magnetic,putzky2019rim,quan2018compressed,radford2015unsupervised,kazeminia2020gans,sarasaen2021fine}. The Variational Network (VarNet) \cite{hammernik2018learning} solves an optimization problem by a cascaded design of multiple layers, with each layer solving a single gradient update step. The End-to-End Variational Network (E2E-VarNet) approach extends the VarNet model by including an estimation of the sensitivity maps within the network, which significantly improves the quality of the reconstruction at higher acceleration rates. Currently, the E2E-VarNet provides state-of-the-art reconstruction performance on the fastMRI data \cite{sriram2020end}.

DL-based models, however, commonly provide the best point estimate of network parameters and do not enable uncertainty quantification which is critical for clinical decision-making \cite{shaul2020subsampled,edupuganti2020uncertainty,hu2021learning,narnhofer2021bayesian,chung2022score,mardani2018deep,meyer2021uncertainty}. They lack provable robustness, and their failure is not fully understood, which raises concerns about their use in real-world scenarios \cite{zhang2023uncertainty,meyer2021uncertainty}. Furthermore, current state-of-the-art DL-based reconstruction methods are susceptible to distribution shifts, such as changes in imaged anatomy or undersampling patterns used for acquisition \cite{morshuis2022adversarial,meyer2021uncertainty,zhang2023uncertainty}.

Monte Carlo dropout approaches \cite{gal2016dropout,mosegaard1995monte}, are widely used in DL-based MRI reconstruction to offer probabilistic interpretability and uncertainty quantification via sampling the predictions \cite{luo2020mri}. However, they are limited to DNN-based models that were trained with dropout layers. Diffusion models have been combined with feed-forward DL models to perform image reconstruction with quantification of uncertainty \cite{chung2022come,chung2022score}. These models, however, incorporate noise in the image space rather than in the physically relevant k-space, which may lead to a loss of anatomical content in the resulting reconstruction.  
Additionally, Variational Autoencoders (VAEs) \cite{cai2019multi,larsen2016autoencoding} are widely used to incorporate a probabilistic perspective to DNNs by assuming that the latent space adheres to a particular probability distribution, typically a multivariate Gaussian \cite{edupuganti2020uncertainty,tezcan2018mr,narnhofer2021bayesian,kazeminia2020gans}. The optimization process in VAEs involves minimizing the Kullback-Leibler divergence between the approximated posterior and the assumed target distribution.



In a shift of these parametric approaches, our study presents a novel non-parametric Bayesian deep neural network (DNN)-based method for magnetic resonance imaging (MRI) image reconstruction from under-sampled k-space data. This approach allows for the full characterization of the posterior distribution of the reconstructed MRI images, providing quantitative measures of uncertainty for the prediction. Our method employs Stochastic Gradient Langevin Dynamics (SGLD) \cite{welling2011bayesian} to sample from the posterior distribution of the network parameters \cite{cheng2019bayesian}, using Gaussian noise injected into the loss gradients during the training process to enable sampling. After a ``burn-in`` iteration where the training loss curve stabilizes, we save the models' parameters for later use in estimating the statistics of the reconstructed image during inference. This method is not restricted to a specific network architecture and can be integrated into existing models. Differing from VAEs, the SGLD method captures the posterior distribution of network parameters by adding noise to the gradients during training. This process does not depend on predefined assumptions about the parameters' distribution or its parametric form. Instead, it enables the determination of the network parameters' posterior by sampling, with the only assumption being that the added noise is normally distributed.


Our main contributions are: 
\begin{itemize}
    \item A non-parametric Bayesian deep-learning method for MRI reconstruction from under-sampled k-space data.
    \item Full characterization of the posterior distributions of the reconstructed images.
    \item Uncertainty estimate of the reconstructed images that correlate with the reconstruction error, anatomical distribution shift, and undersampling mask distribution shift.
    \item Experimental demonstration of the added value of our non-parametric approach over baseline methods.
\end{itemize}

We empirically evaluated the aforementioned hypotheses through experiments utilizing the publicly available fastMRI dataset\footnote{\url{https://fastmri.org/}}, with the E2E-VarNet model as the system's backbone. It is worth highlighting that the original  E2E-VarNet model, serving as the primary framework for this research, lacks capabilities in posterior sampling and uncertainty quantification.


This paper expands upon our preliminary research presented at the International Workshop on Machine Learning for Medical Image Reconstruction, MICCAI \cite{khawaled2022npbdreg}. The current work provides a more comprehensive exposition of the approach and examines the generalization capacity of the method. To this end, we conducted further experiments using both knee and brain data to assess the generalization performance of the approach against anatomical distribution shifts. The earlier work was limited to training and evaluation on the brain dataset only.



\begin{figure*}[t]
\centering
\scalebox{0.8}{	
\centering{
		\pgfdeclarelayer{background}
		\pgfdeclarelayer{foreground}
		\pgfsetlayers{background,main,foreground}
		
		\tikzstyle{sensor}=[draw, fill=gray!30, text width=5em, 
		text centered, minimum height=3em]
		\tikzstyle{ann} = [above, text width=4em]
		\tikzstyle{naveqs} = [sensor, text width=5em, fill=blue!10, 
		minimum height=8em, rounded corners]
		\tikzstyle{naveqs1} = [sensor, text width=6em, fill=red!10, 
		minimum height=11em, rounded corners]
  		\tikzstyle{naveqs2} = [sensor, text width=4em, fill=green!10, 
		minimum height=5em, rounded corners]
		
		\def\blockdist{2.3}
		\def\edgedist{3}
		
		\begin{tikzpicture}
		\node (naveq) at (0,0) [naveqs] {UNet};
		\node (naveq1) at (0.03,0.03) [naveqs] {Rec. Nets};
		\node (naveq2) at (0.13,0.13) [naveqs] {Rec. Nets};
		\node (naveq3) at (0.23,0.23) [naveqs] {Rec. Nets};
		\node (naveq4) at (0.33,0.33) [naveqs] {Rec. Nets};
		\node (naveq41) at (4.3,-0.2) [naveqs1,rotate=0] {Calculate Statistics};
      \node (naveq42) at (0,-3) [naveqs2,rotate=0] {SGLD based Training };
        \node[inner sep=0pt] (kinput) at (-1.4*\blockdist,0.3){\includegraphics[width=.11\textwidth]{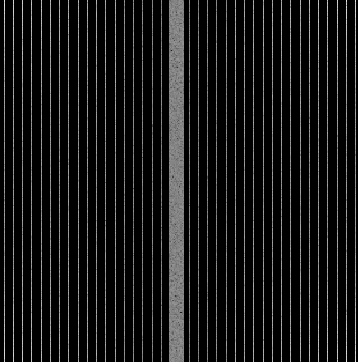}};
        \node[inner sep=0pt] (kinput2) at (-1.5*\blockdist,0.2){\includegraphics[width=.11\textwidth]{mask_eq_R8.png}};
        \node[inner sep=0pt] (recout) at (7.8,0.9){\includegraphics[width=.11\textwidth]{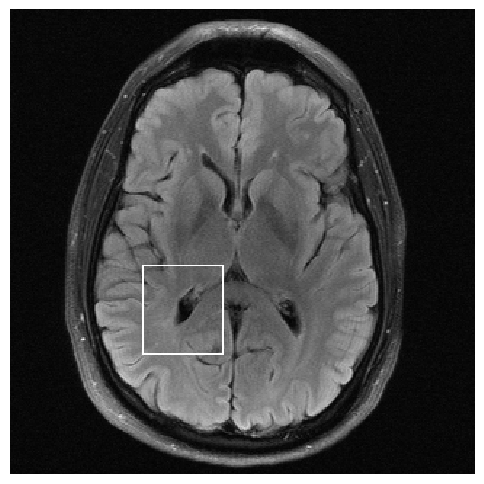}};
         \node[inner sep=0pt] (sigma) at (7.8,-0.9){\includegraphics[width=.11\textwidth]{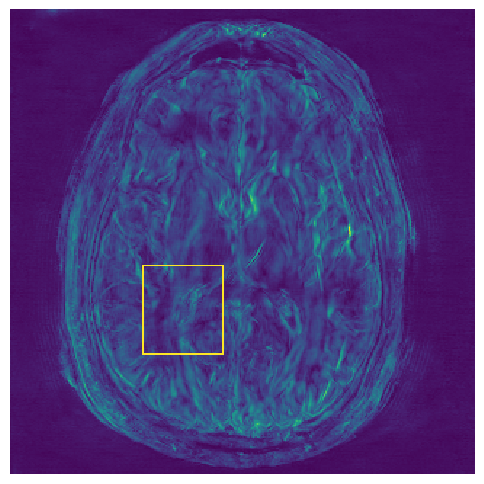}};
         \node[inner sep=0pt] (target) at (-1.4*\blockdist,-3){\includegraphics[width=.11\textwidth]{target_R8.png}};

		
		\path [draw, ->] (kinput) -- node [above] {$\{\tilde{k_i}\}_{i=1}^{N_c}$} 
		(naveq.west |- kinput.east) ;
		\path (naveq.south west)+(-0.6,-0.4) node (INS) {};
		\draw[-] (naveq4) -- node [above](\edgedist,0) {{$\left\{ \hat{x}_{t}\right\} _{t_{b}}^{N}$}} ( naveq41.west |- naveq4);
    	\path [draw, <-] (recout) -- node [above] {$\bar{x}$} 
		(naveq41.east |- recout) ;
		 \path [draw, <-] (sigma) -- node [above] {$\Sigma$} 
		(naveq41.east |- sigma) ;
        \path [draw, ->] (target) -- node [above] {$x$} 
		(naveq42.west |- target.east) ;
          \path [draw, ->] (naveq42) -- node [above] {}(naveq42.north |- naveq.south) ;
         \draw[-] (-1.5,0.3) |- (-1.5,-2.5);
       \draw[->]  (-1.5,-2.5) |- (-0.8,-2.5) ;
  
		\end{tikzpicture}
        }   
        }
        
	\caption{Schematic illustration of our proposed NPB-REC system. At the inference phase, we use a set of models with the same backbone network but different parameters, which resulted from the SGLD-based training. Then, these models predict a set of reconstructed images, $\left\{ \hat{x}_{t}\right\} _{t_{b}}^{N}$ by propagating the under-sampled k-space input data,$\{\tilde{k_i}\}_{i=1}^{N_c}$, through each one of the backbone models. 
	Lastly, the averaged reconstructed image and the pixel-wise std., $\bar{x}$ and $\Sigma$, are calculated. The average is used as the most probable reconstruction prediction and the $\Sigma$ is utilized for uncertainty assessment. }
	\label{fig:bddiagram}
\end{figure*}

\section{Background}
\subsection{MRI Reconstruction from Under-Sampled Data} 
In multi-coil acquisition, the MRI scanner consists of multiple receiver coils, where each of them partially acquires measurements in the k-space (frequency domain) \cite{majumdar_2015}. Each coil modulates the k-space samples according to its spatial sensitivity map to the MRI signal:
\begin{equation}
k_i=S_i\mathcal{F}\left(x\right)+\epsilon_i \:\: \forall i\in[1,..,N_c]    
\end{equation}
where $\mathcal{F}$ is the Fourier transform, $S_i$ denotes the corresponding sensitivity map and $N_c$ is the overall number of coils.
To accelerate the acquisition time, k-space measurements are under-sampled by selecting only a specific subset of the entire k-space data, $\tilde{k_i}=M \circ k_i$, where $M$ is the corresponding binary mask that encodes the under-sampling operator. Restoration of the MRI image from the under-sampled data by performing an inverse Fourier transform on the under-sampled data leads to aliasing artifacts.
\subsection{Deep Learning for MRI Reconstruction}
Recently, several DNN-based methods have been developed to tackle the reconstruction task \cite{shaul2020subsampled,edupuganti2020uncertainty,eo2018kiki,akccakaya2019scan,tezcan2018mr,morris2018magnetic}.
Such models predict the high-quality reconstructed output, $\hat{x}$ through a forward pass of the DNN model:
\begin{equation}
\hat{x} = f_{\theta}(\mathcal{I};D) \label{eq:pred}
\end{equation}
where $\mathcal{I}$ is the input of the network, which is either the undersampled k-space data (single or multi-coil) or its corresponding low-quality representation in the image space, $D$ is the training dataset consisting of pairs of high-resolution images and their corresponding undersampled version, $\theta$ denotes the network parameters, and $f_{\theta}$ denotes a forward pass of the network with the parameters $\theta$.
Eq. \eqref{eq:pred} can be either interpreted as a prediction of fully-sampled k-space data from the under-sampled k-space data \cite{akccakaya2019scan} or as a combination between predictions in the image-space and the k-space \cite{eo2018kiki}. Such networks can be trained in a conventional supervised training process to find the parameters $\hat{\theta}$: 
\begin{equation}
\hat{\theta} = \argmin_{\theta} L(D;\theta)  
\label{eq:opt1}
\end{equation}
Alternatively, unsupervised or generative adversarial approaches can be used to overcome the need for corresponding pairs of inputs and outputs \cite{shaul2020subsampled}.

Variational Network (VarNet) \cite{hammernik2018learning} solves \eqref{eq:opt1} by a cascaded design of multiple layers, each one solves a single gradient update step. A more recent work, the End-to-End Variational Networks (E2E-VarNet) approach \cite{sriram2020end}, extends the VarNet model by including estimation of the sensitivity maps within the network, which, in turn significantly improves the quality of the reconstruction at higher accelerations. 

These DL-based methods solve the optimization problem, formulated as in \eqref{eq:opt1}, by updating the network parameters:
\begin{equation}
    \theta^{t+1}  = \theta^t - \epsilon^t \nabla_{\theta}L(D;\theta) \label{eq:sgdupdate}
\end{equation}

where $\epsilon^t$ is the optimizer step size. However, this process provides a best point estimate of network parameters, $\theta$, rather than characterizing the entire posterior distribution of the network parameters, which doesn't provide an uncertainty measure for the reconstructed image. 
\subsection{A Bayesian Perspective on DNN-based MRI Reconstruction}
Bayesian MRI reconstruction aims to provide an uncertainty estimate by characterizing  the posterior distribution, $P\left(\hat{x}|\mathcal{I},\mathcal{D}\right)$, of the predicted reconstruction $\hat{x}$, given under-sampled k-space input $\mathcal{I}$ and training 
data $\mathcal{D}$, by:
\begin{equation}
P\left(\hat{x}|\mathcal{I},\mathcal{D}\right) = \int_{\Theta} P\left(\hat{x}|\mathcal{I},\theta\right)P\left(\theta|\mathcal{D}\right)\diff\theta \label{eq:4}
\end{equation} 
 where $P\left(\theta|\mathcal{D}\right)$ denotes the posterior distribution of the network parameters $\theta$, given the training data $D$.  
 
The aforementioned integral in \eqref{eq:4} is intractable, however an empirical estimate can be simply computed by averaging $\mathcal{N}$ sets of network parameters, $\left\{\theta_i\right\}_{i=0}^{\mathcal{N}-1}\sim P\left(\theta|\mathcal{D}\right)$:
\begin{equation}
P\left(\hat{x}|\mathcal{I},\mathcal{D}\right)  \simeq \frac{1}{\mathcal{N}} \sum_{\theta_i \sim P\left(\theta|\mathcal{D}\right)} P\left(\hat{x}|\mathcal{I},\theta_i\right) \label{eq:avgpost}
\end{equation}

 The sampling process enables reconstructing the image by averaging predictions, $f_{\theta}(\mathcal{I})$, from the reconstruction network with the different sets of parameters: 
 \begin{equation}
     \hat{x}^* = \frac{1}{\mathcal{N}} \sum_{\theta_i \sim P\left(\theta|\mathcal{D}\right)} f_{\theta_i}(\mathcal{I})\label{eq:recstd} 
 \end{equation} 
 where $\hat{x}^*$ is the average of the predicted high-quality images given the input $\mathcal{I}$. Similarly, the variance of the reconstructed image is given by:
 \begin{equation}
     \hat{\sigma_x}^2 = \frac{1}{\mathcal{N}-1} \sum_{\theta_i \sim P\left(\theta|\mathcal{D}\right)} (f_{\theta_i}(\mathcal{I})-\hat{x}^*)^2 \label{eq:recvar}
 \end{equation} 
The latter is used to quantify the uncertainty of the reconstruction. 
Supplementary techniques such as Markov Chain Monte Carlo (MCMC) \cite{robert1999monte} facilitate the characterization of the posterior of the DNNs parameters $\left\{\theta_i\right\}_{i=0}^{\mathcal{N}-1}\sim P\left(\theta|\mathcal{D}\right)$ through sampling. However, convergence with MCMC techniques is generally slower than backpropagation for DNN-based models \cite{neal2012bayesian}.
The Monte Carlo dropout approach \cite{luo2020mri} is widely employed to facilitate the integration of MCMC samplers into DNN-based reconstruction models. However, it is confined to specific DNN-based architectures and can only be applied to models that have been trained with dropout layers.

\subsection{Stochastic Gradient Langevin Dynamics} 
Stochastic gradient Langevin dynamics (SGLD), proposed by Welling and Teh \cite{welling2011bayesian}, provides a general framework to efficiently derive an MCMC sampler
from Stochastic gradient descent (SGD) by injecting Gaussian noise into the gradient updates. The standard update in \eqref{eq:sgdupdate} turn into the following SGLD update \cite{welling2011bayesian}: 
\begin{equation}
    \theta^{t+1} = \theta^t -\epsilon^t \nabla_{\theta}L(\mathcal{I};D,\theta) + \textbf{N}^{t} \label{eq:sgldupdate}
\end{equation}
where $\textbf{N}^{t}\sim\mathcal{N}(0,\epsilon^t)$ and $\epsilon^t$ is the step size. Under suitable conditions ($\sum_t \epsilon^t = \infty$ and $\sum_t (\epsilon^t)^2 < \infty$), the distribution of $\theta$ converges to the true posterior $P\left(\theta|\mathcal{D}\right)$, expressed in \eqref{eq:4}, see \cite{welling2011bayesian} for SGLD convergence
rate proofs and analysis. In advanced stages of the optimization, the impact of the gradient shrinks and the Gaussian noise becomes the dominant
term, transitioning from SGD to Langevin Monte
Carlo \cite{neal2012bayesian}. 

Recently, the use of SGLD as MCMC sampler in DNN, which enables characterization of the posterior, was proposed by \cite{cheng2019bayesian,li2016preconditioned} and further extended to the \texttt{Adam} optimizer by \cite{neelakantan2015adding}.   
This mechanism guides the optimizer to provide a characterization of the posterior distribution rather than finding the maximum a posteriori (MAP) solution without making any specific assumption on $P\left(\theta\right)$ \cite{welling2011bayesian}. 


Consequently, the reconstructed image can be calculated simply by averaging the images reconstructed with the different samples of the network parameters, avoiding the need to integrate the posterior as in ~\eqref{eq:4}. The uncertainty measure can be assessed through the variance of these samples, given by ~\eqref{eq:recvar}.

\section{Non-parametric Bayesian Approach for MRI Reconstruction}\label{subsec:offlinetraining}
 In this work, we consider an SGLD sampler to generate the posterior samples of the reconstructed image. To this end, we incorporate a noise scheduler that injects a time-dependent Gaussian noise to the gradients of the loss during the optimization process with an \texttt{Adam} optimizer.  
\paragraph{Offline Training}
Let $L^{t}$ be the overall reconstruction loss at the training iteration (epoch) $t$, which is calculated between the network predictions $\hat{x}=f_\theta\left(\mathcal{I}\right)$ and the ground truth images, $x$. The most frequently used loss functions are: mean squared error (MSE), mean absolute error (MAE), structural similarity (SSIM) \cite{wang2004image}, etc. 
Then, the loss gradients are denoted by:
\begin{equation}
g^t\overset{\triangle}{=}\nabla_{\theta}L^t\left(x,f_\theta\left(\mathcal{I}\right)\right)
\end{equation}
At every training iteration, $t$, we add Gaussian noise to $g$:
\begin{equation}
\tilde{g}^{t}\leftarrow g^{t}+\textbf{N}^{t}
\end{equation}
where $\textbf{N}^{t}\sim\mathcal{N}(0,s^t)$, $s^t$ is a user-selected parameter that controls the noise variance (can be time-decaying or a constant). Inspired by \cite{neelakantan2015adding}, we selected $s^t$ equal to the \texttt{Adam} learning rate, to avoid escaping from the achieved minimum when the loss is stable. Adding large-magnitude noise may disrupt the trajectory of the optimization process, causing significant perturbations to the model parameters during each iteration. 


Afterwards, the network parameters are updated in the next iteration according to the ``noisy'' gradients: 
according to the \texttt{Adam} update rule:
\begin{equation} 
\theta^{t+1}\leftarrow \theta^{t} - s^{t}\hat{m}^{t}
\end{equation}
where $s^{t}=\frac{\eta}{\sqrt{\hat{v}^{t}+\epsilon}}$, $\hat{m}^{t}$ and $\hat{v}^{t}$ are the bias-corrected versions of the decaying averages of the past gradients and the second moment (squared gradients), respectively:
$\hat{m}^{t} = \nicefrac{m^{t}}{1-\beta_{1}^{t}}$
, $\hat{v}^{t} = \nicefrac{v^{t}}{1-\beta_{2}^{t}}$,
where $m^{t}=m^{t-1}+(1-\beta_{1})\tilde{g}^{t}$ and $v^{t}=v^{t-1}+(1-\beta_{2})(\tilde{g}^{t})^2$. $\beta_{1}$, $\beta_{2}$ are decay rates and $\eta$ is a fixed constant.  
This noise schedule can be performed with any stochastic optimization algorithm during the training procedure and is not limited to \texttt{Adam} update rule only.

Lastly, we save the sets of network parameters $\left\{\theta_i\right\}_{i=t_b}^{\mathcal{N}-1}$ that were obtained in iterations $t\in \left[t_b,\mathcal{N}\right]$, where $\mathcal{N}$ is the overall number of iterations and $t_b$ is the SGLD-parameter. $t_b$ should be larger than the cut-off point of the \textit{burn-in} phase. One should sample parameters obtained in the last $t_b,..,\mathcal{N}$ iterations, where the loss curve has converged. 
\paragraph{Inference time}
We exploit the network parameters that were obtained after the \textit{burn-in} phase, i.e. in the last $t_b,..,\mathcal{N}$ iterations, $\left\{ \theta\right\}_{t_{b}}^{\mathcal{N}}$ to generate a set of reconstructed images from which we can calculate the average image and its standard deviation as an uncertainty measure. Fig.~\ref{fig:bddiagram} illustrates the operation of the NPB-REC system at the inference phase. The parameter $\mathcal{N}-t_b$, i.e the number of saved models, is selected after performing hyper-parameters selection on a subset of the evaluation data, which will be addressed in detail in subsection \ref{subsec:hypertuning}
We sample a set of reconstructed images  $\left\{ \hat{x}\right\}_{t_{b}}^{\mathcal{N}}$, obtained by feed-forwarding the input data $\mathcal{I}$ to the reconstruction models with the parameters $\left\{ \theta\right\}_{t_{b}}^{\mathcal{N}}$.
Then, we estimate the averaged posterior image to achieve a high-quality reconstruction: 
\begin{equation}
\hat{x}^{*}=\frac{\sum_{t=t_b}^{\mathcal{N}} \hat{x}_t}{\mathcal{N}-t_b} \label{eq:avg}
\end{equation} 
In addition, we quantify the standard deviation of the reconstructed image \eqref{eq:recstd}, which is used to characterize the uncertainty.

\paragraph{The Backbone Reconstruction Network}
The backbone of our reconstruction system is based on the E2E-VarNet model \cite{sriram2020end}. E2E-Varnet contains multiple cascaded layers, each applying a refinement step in the k-space according to the following update: 
\begin{equation}
    k^{m+1} = k^m - \eta^t M \left(k^m-\tilde{k}\right)+G\left(k^m\right) \label{eq:kspacesolve}
\end{equation} 
where $k^m$ and $k^{m+1}$ are the input and output to the m-th layer, respectively. $G$ is the refinement module: $ G(k^m) = \mathcal{F}\circ\mathcal{E}\circ \text{CNN}(\mathcal{R}\circ\mathcal{F}^{-1}(k^m))$.
 The CNN maps a complex image input to a complex output, $\mathcal{E}$ and $\mathcal{R}$ are the expand and reduce operators. $\mathcal{E}$ computes the corresponding image seen by each coil: $\mathcal{E}(x) = (S_{1}x,...,S_{N_c}x)$ and its inverse operator, $\mathcal{R}$, integrates the multiple coil images $\mathcal{R}(x_1,...,x_{N_c})=\sum S_ix_i$. 
Similarly to the design of E2E-VarNet \cite{sriram2020end}, a U-Net is used as the CNN \cite{ronneberger2015u}. Additional U-Net with the same architecture of CNN is used to estimate the sensitivity maps $S_1,...,S_{N_c}$ during the reconstruction. After applying the cascaded layers to the k-space input, as described in \eqref{eq:kspacesolve}, we obtain the final reconstructed image, $\hat{x}$,  by root-sum-squares (RSS) reduction of the image-space representation of the final layer output: $\hat{x} = \sqrt{\sum_{i=1}^{N_c}|\mathcal{F}^{-1}k^{T}_i|^2}$.

\begin{figure}[t!]
\centering
\includegraphics[width=7cm]{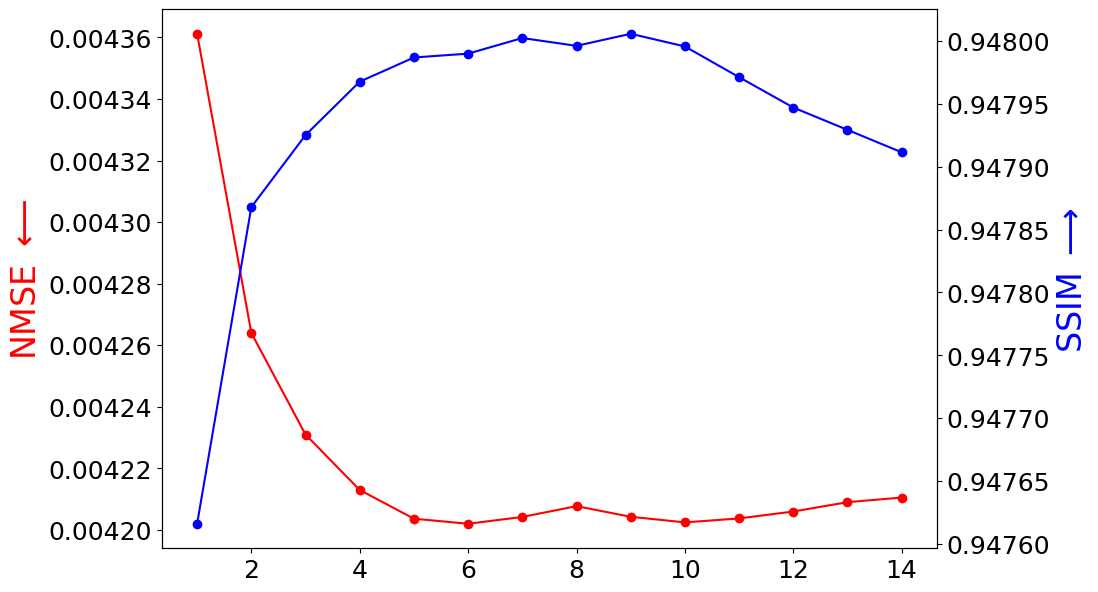}
\caption{Normalized MSE (NMSE) and SSIM  vs. $N-t_b$, the number of models used in the averaging, obtained on a subset from the test set (32 images sampled randomly). Lower NMSE curves correspond to better prediction quality. However, a higher SSIM is better. See \textcolor{red}{red} and \textcolor{blue}{blue} arrows.}\label{fig:tune}
\end{figure} 

\begin{figure*}[t!]
\centering
	\begin{minipage}[b]{0.18\linewidth}
		\centering
		\includegraphics[width=2.6cm]{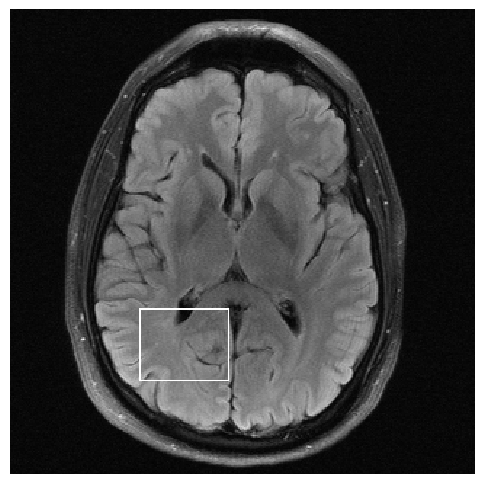}
	\end{minipage} 
	\begin{minipage}[b]{0.18\linewidth}
		\centering
		\includegraphics[width=2.6cm]{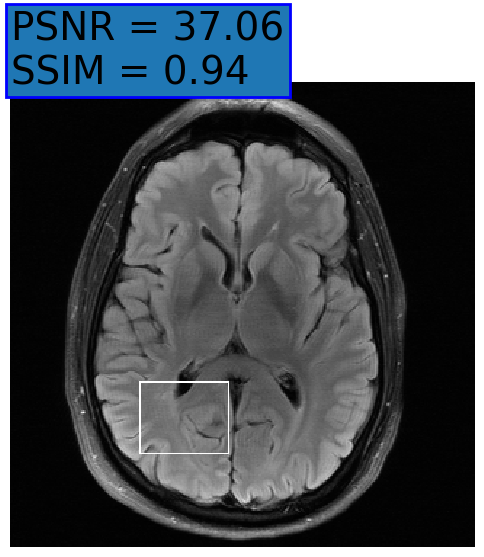}
	\end{minipage}
	\begin{minipage}[b]{0.18\linewidth}
		\centering
		\includegraphics[width=2.6cm]{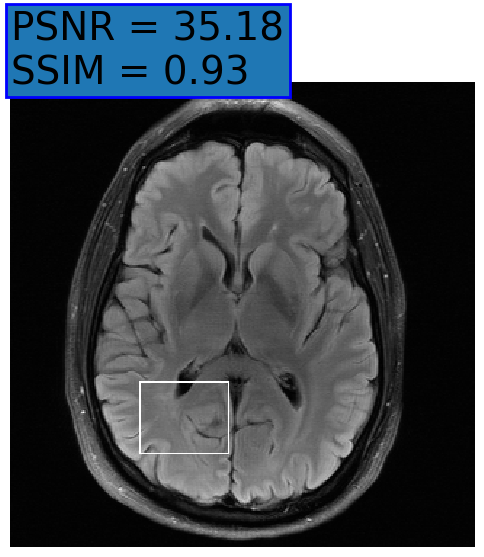}
	\end{minipage} 
	\begin{minipage}[b]{0.18\linewidth}
		\centering
		\includegraphics[width=2.6cm]{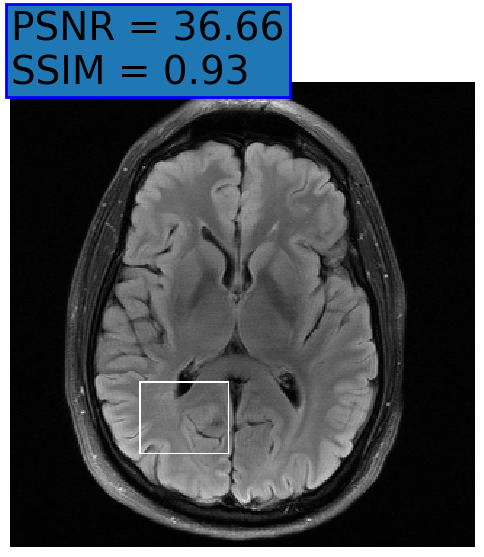}
	\end{minipage} 
		\begin{minipage}[b]{0.18\linewidth}
		\centering
		\includegraphics[width=2.6cm]{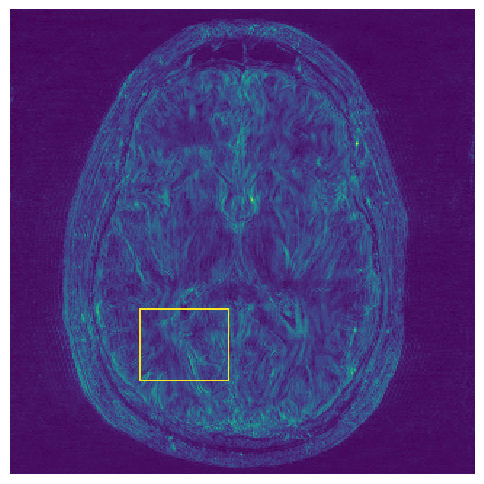}
	\end{minipage}
	\\
	\begin{minipage}[b]{0.18\linewidth}
		\centering
		\includegraphics[width=2.6cm]{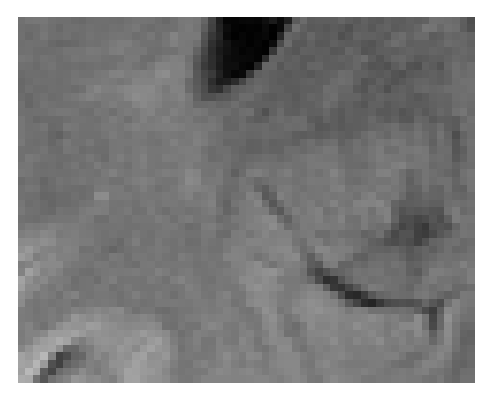}
	\end{minipage} 
	\begin{minipage}[b]{0.18\linewidth}
		\centering
		\includegraphics[width=2.6cm]{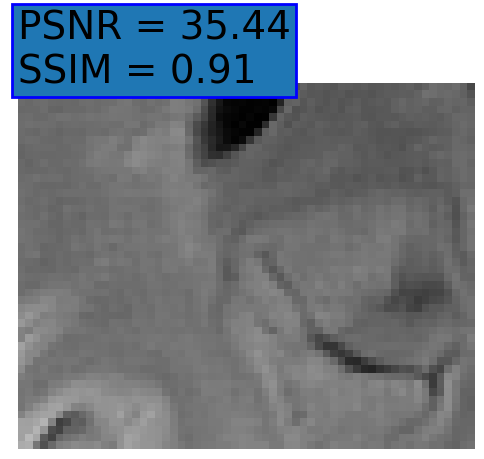}
	\end{minipage} 
		\begin{minipage}[b]{0.18\linewidth}
		\centering
		\includegraphics[width=2.6cm]{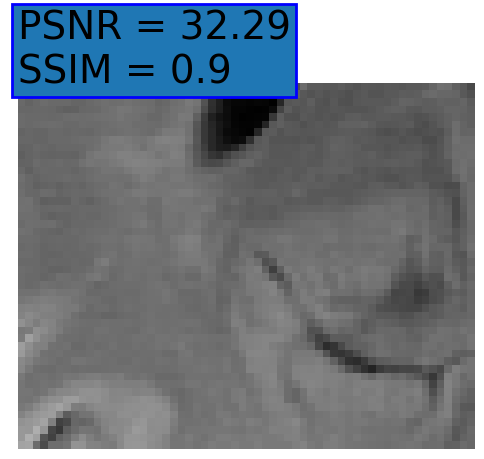}
	\end{minipage} 
		\begin{minipage}[b]{0.18\linewidth}
		\centering
		\includegraphics[width=2.6cm]{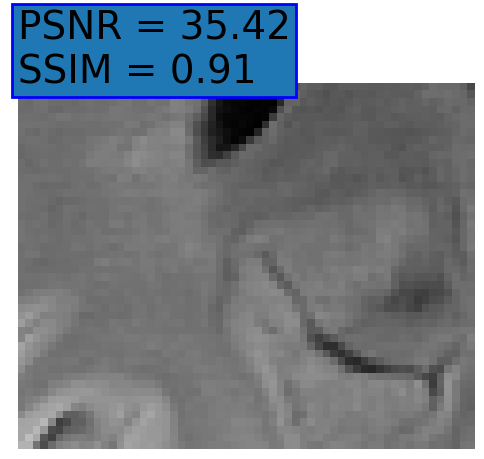}
	\end{minipage} 
		\begin{minipage}[b]{0.18\linewidth}
		\centering
		\includegraphics[width=2.6cm]{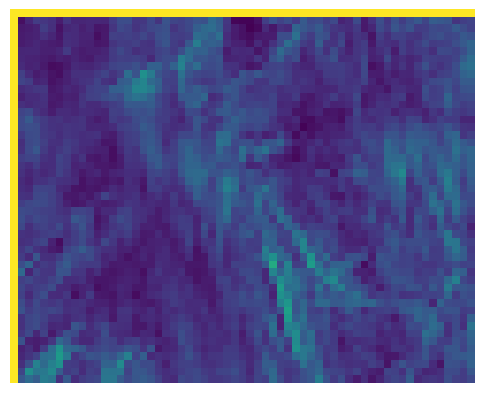}
	\end{minipage}
    \\
	\begin{minipage}[b]{0.18\linewidth}
		\centering
		\includegraphics[width=2.6cm]{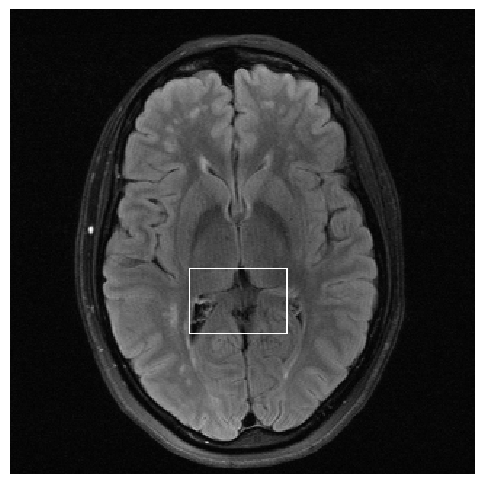}
	\end{minipage} 
	\begin{minipage}[b]{0.18\linewidth}
		\centering
		\includegraphics[width=2.6cm]{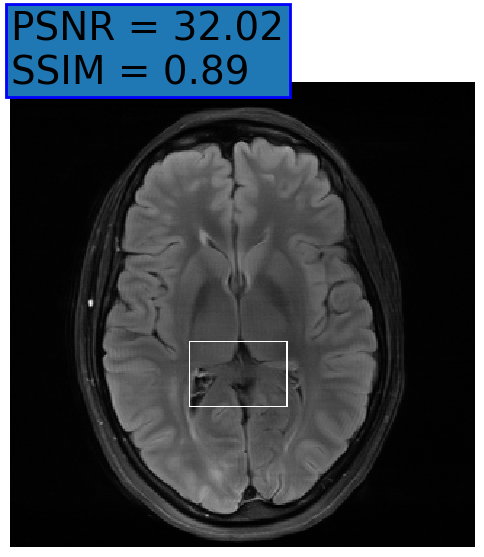}
	\end{minipage}
	\begin{minipage}[b]{0.18\linewidth}
		\centering
		\includegraphics[width=2.6cm]{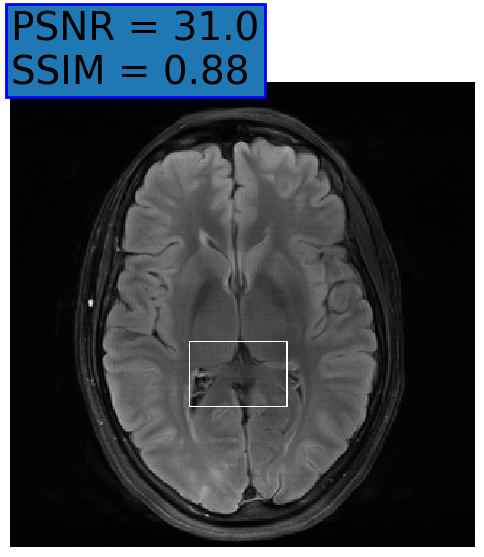}
	\end{minipage} 
	\begin{minipage}[b]{0.18\linewidth}
		\centering
		\includegraphics[width=2.6cm]{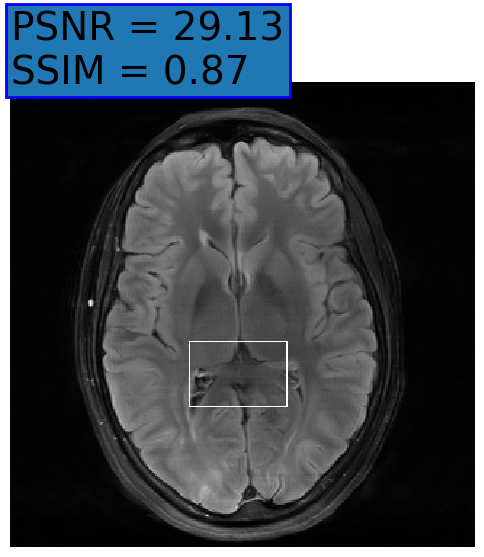}
	\end{minipage} 
		\begin{minipage}[b]{0.18\linewidth}
		\centering
		\includegraphics[width=2.6cm]{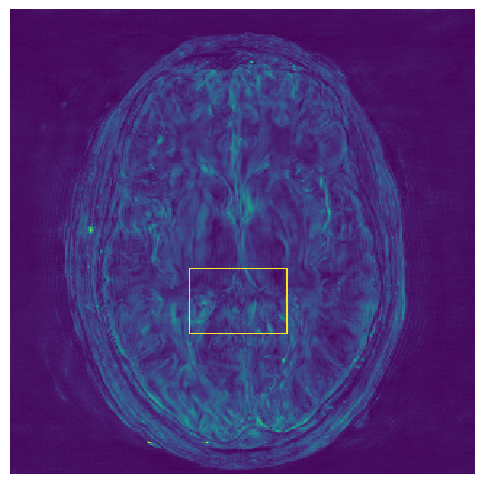}
	\end{minipage}
	\\
		\begin{minipage}[b]{0.18\linewidth}
		\centering
		\subfloat[\label{fig2:a}\footnotesize GT]{\includegraphics[width=2.6cm]{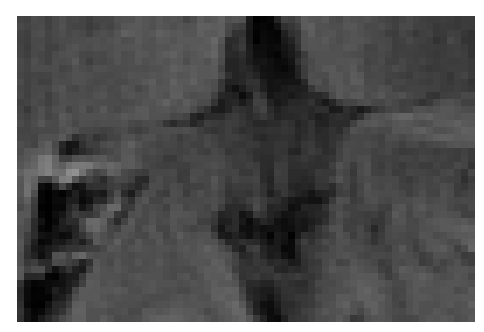}}
	\end{minipage} 
	\begin{minipage}[b]{0.18\linewidth}
		\centering
		\subfloat[\label{fig2:b}\footnotesize NPB-REC Avg.]{\includegraphics[width=2.6cm]{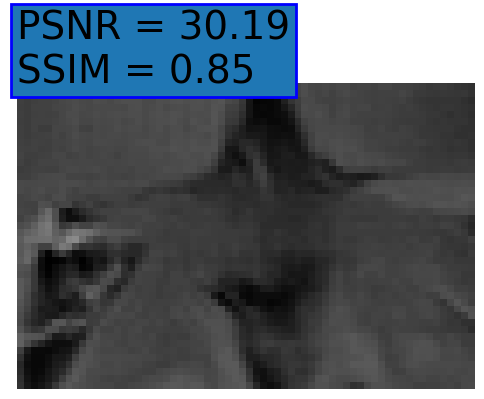}}
	\end{minipage} 
		\begin{minipage}[b]{0.18\linewidth}
		\centering
		\subfloat[\label{fig2:c}\footnotesize baseline]{\includegraphics[width=2.6cm]{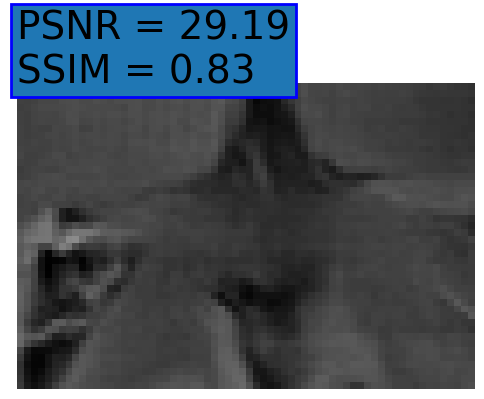}}
	\end{minipage} 
		\begin{minipage}[b]{0.18\linewidth}
		\centering
		\subfloat[\label{fig2:d}\footnotesize Dropout]{\includegraphics[width=2.6cm]{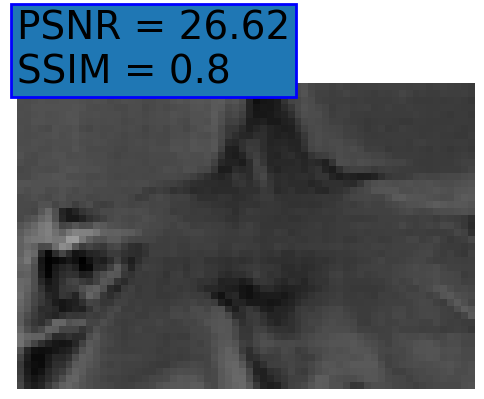}}
	\end{minipage} 
		\begin{minipage}[b]{0.18\linewidth}
		\centering
		\subfloat[\label{fig2:e}\footnotesize NPB-REC Std.]{\includegraphics[width=2.6cm]{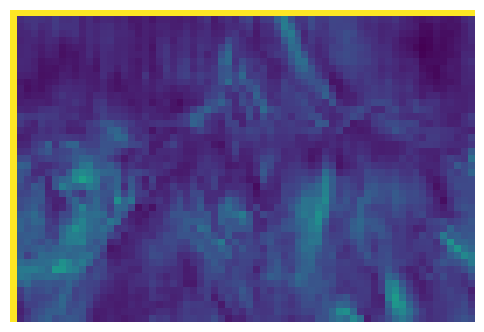}}
	\end{minipage}
	\caption{Reconstructions results. Rows 1 and 3: Examples of comparing the ground truth (GT) fully sampled image to the reconstructed images obtained by the three models (1-3), NPB-REC, baseline, E2E-VarNet trained with Dropout, and the NPB-REC std. map at accelerations $R=4$, $R=8$, respectively. Rows 2 and 4: The corresponding annotated ROIS of Nonspecific white matter lesions. }\label{fig:RecMRIexample}	
\end{figure*} 
\section{Experimental Methodology}
\subsection{Database}
In our experiments, we used multi-coil data of brain and knee MRI images, adapted from the publicly available fastMRI database 
   for training our system; we used the validation and training datasets for the brain and knee multi-coil challenge. We excluded a subset from the validation set that contained clinical pathology annotations taken from fastMRI+ \cite{zhao2021fastmri} and used it in the evaluation of our method, where no overlap between subjects would be in these sets. This is due to the fact that the ground truth of the test set is not publicly available and it is interesting to demonstrate our NPB-REC method on regions of interest (ROIs) and to quantify both the reconstruction accuracy and uncertainty in these ROIs. 
  
For the brain data, the training, validation, and inference sets included $4,469$, $1,131$, and $247$ 3D images cropped to $320\times 320$ and k-space data of $20$ coils. The number of slices is not uniform and varies within scans. From these datasets, we generated $88,832$,  $18,080$ slices, and $3,880$ slices with annotations for the training, validation, and evaluation phases, respectively. It is worth mentioning that each slice has multiple annotation labels. For the knee data, the training, validation, and inference sets included $973$, $44$, and $150$ images cropped to $320\times 320$ and k-space data of $15$ coils. $36,504$, $1,760$ 2D images, and $5,585$ annotated slices were generated for the training, validation, and evaluation phases, respectively. 

We generated under-sampled k-space data from the fully-sampled datasets with two types of Cartesian masks: \textit{equispaced} and \textit{random}, which were used for the fastMRI paper \cite{hammernik2018learning}. The former samples $l$ low-frequency lines from the center of k-space and every R-th line from the higher frequencies, to make the acceleration rate equal to $R$. Whereas, the latter samples a fraction of the full width of k-space for the central k-space corresponding to low frequencies and selects uniformly at a subset of a higher frequency line such that the overall acceleration is $R$. In addition, we generated radially undersampled masks to assess the quality of the reconstructions obtained by our NPB-REC model corresponding to a non-Cartesian trajectory. The radial masks were implemented using the CIRcular Cartesian UnderSampling (CIRCUS) method \cite{liu2014accelerated}. 

\subsection{Implementation Details }\label{subsec:hypertuning}
We trained three models: (1) E2E-VarNet \cite{sriram2020end} trained with the proposed NPB-REC method (section~\ref{subsec:offlinetraining}), (2) The baseline E2E-VarNet model \cite{sriram2020end}, and; (3) E2E-VarNet trained with Dropout of probability $0.001$ and Monte Carlo averaging used at inference. Higher values of Dropout probabilities led to instabilities in the network loss during training. The three aforementioned models had the same architecture as described in \cite{sriram2020end} with $T=8$ cascaded layers. Dropout layers were incorporated to model (3) only, whereas the first two configurations were trained without adding dropout layers. For these three models, we used $L=1-SSIM$ as a training loss and \texttt{Adam} optimizer with its default values of $\beta_{1}$, $\beta_{2}$, $\eta$ and a learning rate of $lr=0.001$. The total number of epochs (training iterations) was set to $40$ and the batch size was equal to $1$, due to memory space limitations. The training procedure did not include any regularization or data augmentation. Training of the three models was implemented using the PyTorch-lightning package \cite{falcon2019pytorch}. We use the default PyTorch initialization for network parameters as in \cite{sriram2020end}.
In our experiments, we selected a standard deviation, $s^t=lr$ for the injected noise variance. 
The network parameters are then updated according to the \texttt{Adam} update rule. During training, we generated under-sampled inputs by multiplying with \textit{random} Cartesian masks of acceleration rate $R=4$, which vary during the training process. 
\paragraph{Hyper-parameters Selection:} The dataset for hyper-parameter optimization included 32 images sampled randomly from the test set. We retrospectively undersampled the k-space data with \textit{random} Cartesian masks of acceleration rate $R=4$ as in the training data. We generated a set of reconstructed images obtained by passing under-sampled k-space inputs to the models with the parameters $\left\{ \theta\right\}_{t_{b}}^{N}$, i.e. obtained in the last $N-t_b$ iterations. $t_b$ was selected such that the training loss in the last $N-t_b$ is stable and has only slight variations around its steady state value. In our experiments, we performed a hyper-parameter tuning on $t_b$ and selected the last $N-t_b$ that obtained the best quantitative reconstruction performance on the hyper-parameter optimization test set. 
 
 Fig.~\ref{fig:tune} presents the normalized mean squared error (NMSE) and SSIM metrics for a range of $N-t_b$ that varies from 1 to 10. The NMSE is calculated by normalizing the Mean Squared Error (MSE) value by the mean value of the reconstructed image.
 Although this range of $N-t_b$ values shows similar NMSE and SSIM metrics, we selected $N-t_b=9$. This is due to the fact that it shows a slight improvement and with having $9$ samples we can calculate robust statistics. 

The final reconstructed image was calculated by averaging these $9$ samples, predicted from our model, as mentioned in \eqref{eq:avg}. 
\begin{table*}[t]
\caption{Reconstruction accuracy. Rows top to bottom: PSNR and SSIM metrics calculated on the annotated anatomical ROIs (denoted by 'A') with a mask of acceleration rate $R=4$, the whole physical images (denoted by 'W') with masks of acceleration rate $R=4$, $R=8$, respectively. 'r' and 'e' stands for \textit{random} and \textit{equispaced} mask types.} \label{tab:results}
\setlength{\tabcolsep}{9pt}
\centering
\resizebox{\textwidth}{!}{%
\begin{tabular}{@{}lcccccccc@{}}
\toprule
&                                                        &                             & \multicolumn{2}{c}{\textbf{NPB-REC}}                                & \multicolumn{2}{c}{\textbf{Baseline}}                                                   & \multicolumn{2}{c}{\textbf{Dropout}}                                                    \\ \midrule
\multicolumn{1}{l}{}                    & \multicolumn{1}{c}{\textbf{R}}                                                                     & \multicolumn{1}{c}{\textbf{M}}   & \multicolumn{1}{c}{\textbf{PSNR}}            & \multicolumn{1}{c}{\textbf{SSIM}}               & \multicolumn{1}{c}{\textbf{PSNR}}            & \multicolumn{1}{c}{\textbf{SSIM}}               & \multicolumn{1}{c}{\textbf{PSNR}}            & \multicolumn{1}{c}{\textbf{SSIM}}               \\ \midrule
\multicolumn{1}{l}{\multirow{2}{*}{A}} & \multicolumn{1}{c}{\multirow{2}{*}{\begin{tabular}[c]{@{}c@{}}4\end{tabular}}} & \multicolumn{1}{c}{r} &  \multicolumn{1}{c}{$\mathbf{30.04\pm6.78}$} & $\mathbf{0.87\pm0.18}$ & \multicolumn{1}{c}{$29.91\pm6.87$} & $0.867\pm0.182$   & \multicolumn{1}{c}{$29.5\pm 6.844$}  & $0.858\pm 0.19$  \\ \cmidrule(l){3-9}  
\multicolumn{1}{l}{}                   & \multicolumn{1}{c}{}                                                                       & \multicolumn{1}{c}{e}   & \multicolumn{1}{c}{$\mathbf{32.22\pm 6.94}$} & $\mathbf{0.914\pm0.143}$  & \multicolumn{1}{c}{$32.02\pm 7.35$} & $0.911\pm0.143$  & \multicolumn{1}{c}{$31.57\pm6.89$} & $0.905\pm0.151$    \\ \midrule
\multicolumn{1}{l}{\multirow{2}{*}{W}} & \multicolumn{1}{c}{\multirow{2}{*}{4}}                      & \multicolumn{1}{c}{r} & \multicolumn{1}{c}{$\mathbf{40.24\pm 6.19}$} & $\mathbf{0.947\pm0.081}$ & \multicolumn{1}{c}{$40.17\pm 6.19$} & $\mathbf{0.947\pm0.081}$  & \multicolumn{1}{c}{$39.86\pm6.10$} & $0.945\pm0.082$ 
\\ \cmidrule(l){3-9} 
\multicolumn{1}{l}{}                   & \multicolumn{1}{c}{}                                                                       & \multicolumn{1}{c}{e}   &  \multicolumn{1}{c}{$41.61\pm6.28$} & $\mathbf{0.955\pm0.073}$ & \multicolumn{1}{c}{$\mathbf{41.64\pm6.28}$} & $\mathbf{0.955\pm0.074}$ & \multicolumn{1}{c}{$41.22\pm6.0$} & $0.953\pm0.074$ 
\\ \midrule
\multicolumn{1}{l}{\multirow{2}{*}{W}} & \multicolumn{1}{c}{\multirow{2}{*}{8}}                                                   & \multicolumn{1}{c}{r} & \multicolumn{1}{c}{$\mathbf{32.23\pm6.63}$} & $\mathbf{0.881\pm0.11}$ & \multicolumn{1}{c}{$31.21\pm6.2$} & $0.87\pm0.108$  & \multicolumn{1}{c}{$30.63\pm5.91$} & $0.865\pm0.11$  
\\ \cmidrule(l){3-9} 
\multicolumn{1}{l}{}                   & \multicolumn{1}{c}{}    & e  & \multicolumn{1}{c}{$\mathbf{34.55\pm5.01}$} & $\mathbf{0.908\pm0.09}$ & \multicolumn{1}{c}{$33.08\pm4.82$} & $0.897\pm0.092$ & \multicolumn{1}{c}{$32.25\pm4.74$} & $0.891\pm0.09$ \\ 
\bottomrule
\end{tabular}
}
\end{table*}

\begin{figure*}[t!]
	\begin{minipage}[b]{0.32\linewidth}
		\centering
		\subfloat[\label{fig4:a}\footnotesize NPB-REC]{
		\includegraphics[width=3.9cm]{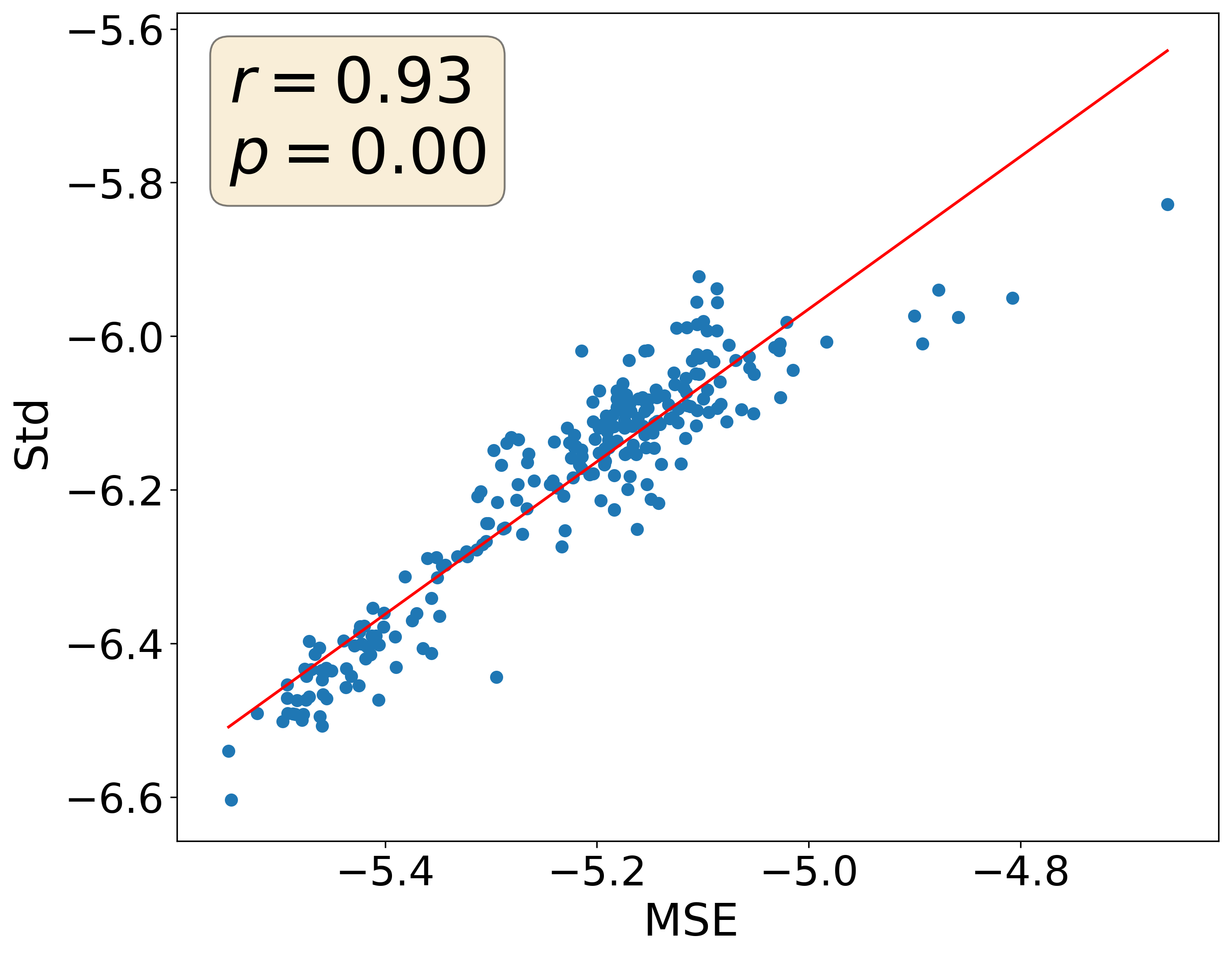}}
	\end{minipage}
	\begin{minipage}[b]{0.32\linewidth}
		\centering
		\subfloat[\label{fig4:b}\footnotesize Dropout]{
		\includegraphics[width=3.9cm]{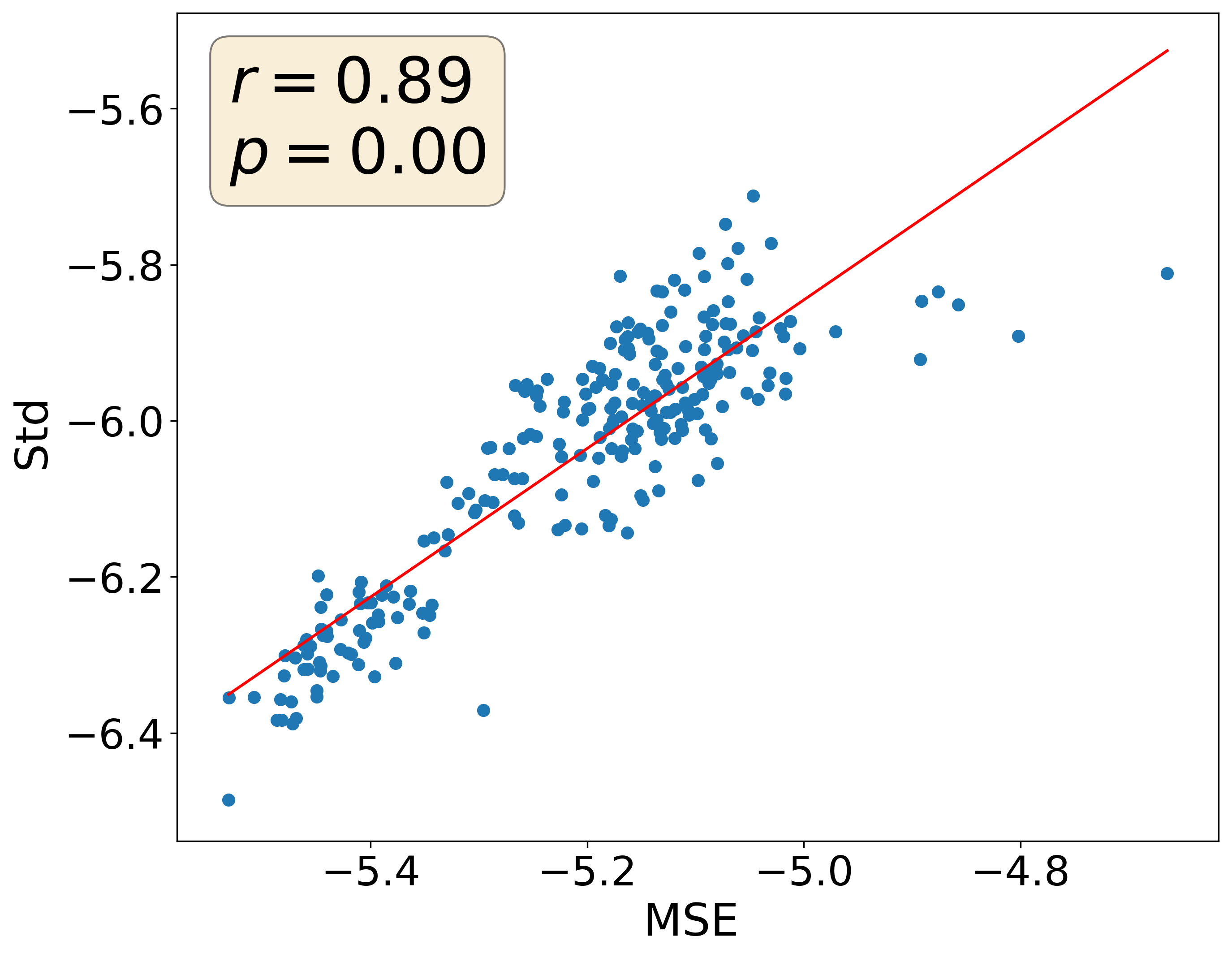}}
	\end{minipage}	
	\begin{minipage}[b]{0.32\linewidth}
		\centering
		\subfloat[\label{fig4:c}  Uncertainty vs. $R$]{
		\includegraphics[width=3.9cm]{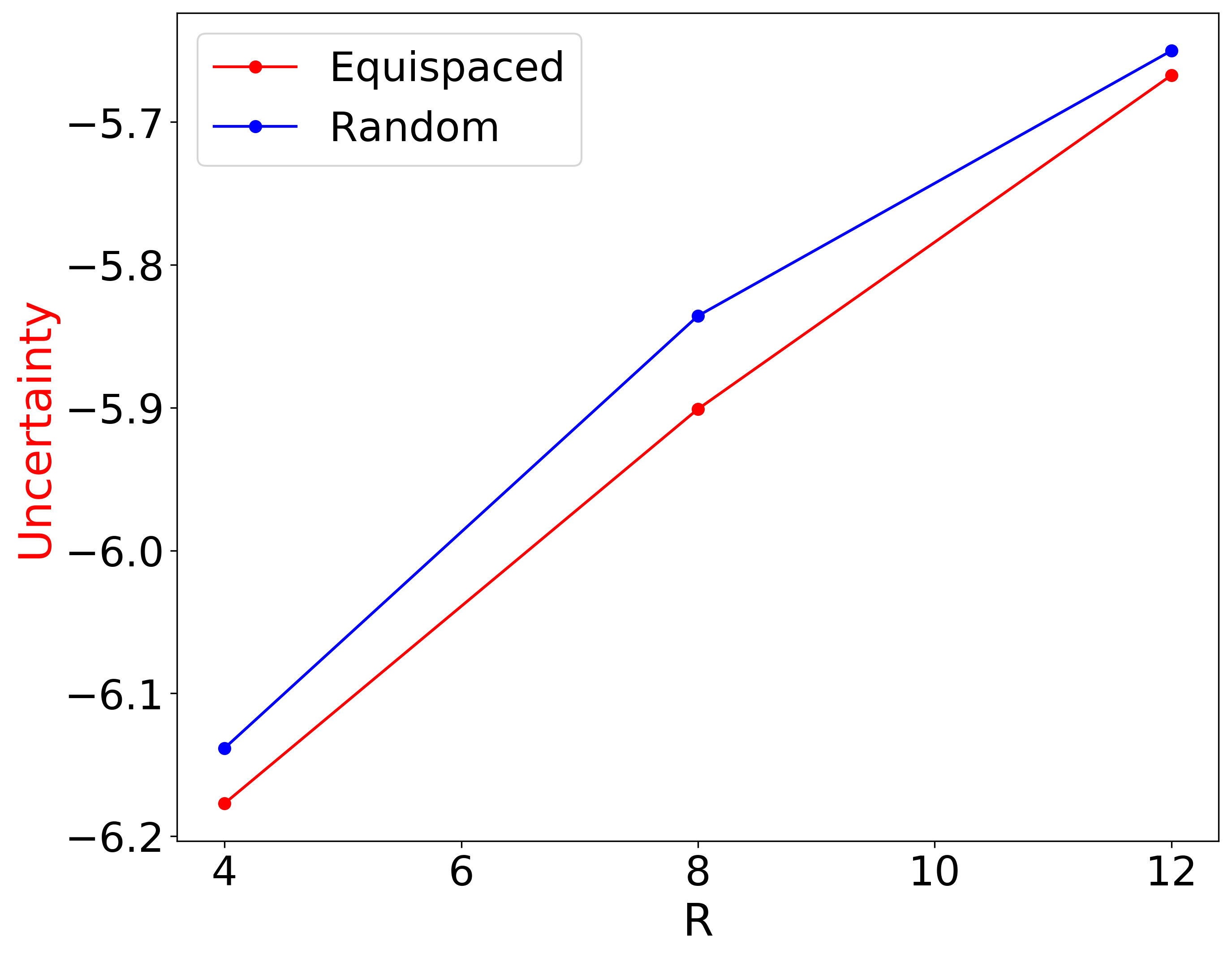}}
	\end{minipage}
	\caption{Uncertainty assessment. Scatter plots of the mean value of Std. estimate versus the MSE metric, calculated between the reconstructed and the ground truth, in log scale, for our NPB-REC method \protect\subref{fig4:a} and Monte Carlo Dropout \protect\subref{fig4:b}. \protect\subref{fig4:c} Our measure of uncertainty versus the acceleration rate.  }\label{fig:unc}
\end{figure*}
\subsection{Reconstruction accuracy}
We first assessed the accuracy of the different reconstruction models. We used test data that was retrospectively undersampled with \textit{random} Cartesian masks of acceleration rate $R=4$ as in the training data. We excluded the 32 images used for hyper-parameter optimization from the test set. We calculated the PSNR and the SSIM between the reconstructed and the ground truth, for all pairs of images in the test set.  
\subsection{Uncertainty Assessment} \label{sebsec:uncasses}
Finally, we assessed the clinical relevance of our uncertainty measures. We estimated the 2D uncertainty maps by calculating the pixel-wise std. of the predictions made by our model. We utilized these pixel-wise std. maps as a measure of uncertainty of the reconstruction.
For comparison, we used the same number of predictions (i.e. $9$ in Monte Carlo sampling at the inference phase, but with enabled Dropout layers. We validated the clinical relevance of our uncertainty measure by inspecting its correlation with the reconstruction error, in comparison with the measurements obtained by Dropout. The reconstruction error used in this scenario is the MSE metric, calculated between the predicted and the ground truth high-resolution image. We also evaluated whether the uncertainty measures can be useful in determining out-of-distribution data by assessing group differences between within-distribution and out-of-distribution data.
\subsection{Improved Generalization Capability}
We then analyzed the generalization capabilities achieved by our NPB-REC approach by means of robustness to both anatomical (brain vs. knee) and sampling pattern distribution shifts (Cartesian vs. non-Cartesian). 
\paragraph{Anatomical Shift}
We first assessed the improved generalization ability of the NPB-REC approach against the benchmarks (2) the baseline and (3) Dropout methods. We assessed the reconstruction performance, by means of SSIM and PSNR, on a test set outside the anatomy of the training data. Specifically, in our experiments, we evaluated the performance of our reconstruction system in MRI images of the knee, while the model was trained on MRI images of the brain only and vice versa.   
\paragraph{Sampling Patterns}
We investigated the ability of our model to generalize when having a sampling pattern distribution shift. To this end, we applied non-Cartesian (radial) masks to generate the test data and then calculated the reconstruction accuracy of NPB-REC compared to the other benchmarks. We used radial masks with two different numbers of lines: $L=50$ and $L=40$, which are equivalent to $R\simeq 6$ and $R\simeq 8$. We performed this experiment only on the annotated brain data, where the models were trained on brain MRI images as well.

\begin{figure}[t!]
	\begin{minipage}[b]{0.23\linewidth}
		\centering
		\subfloat[\label{fig5:a}\footnotesize SGLD]{
		\includegraphics[width=3cm]{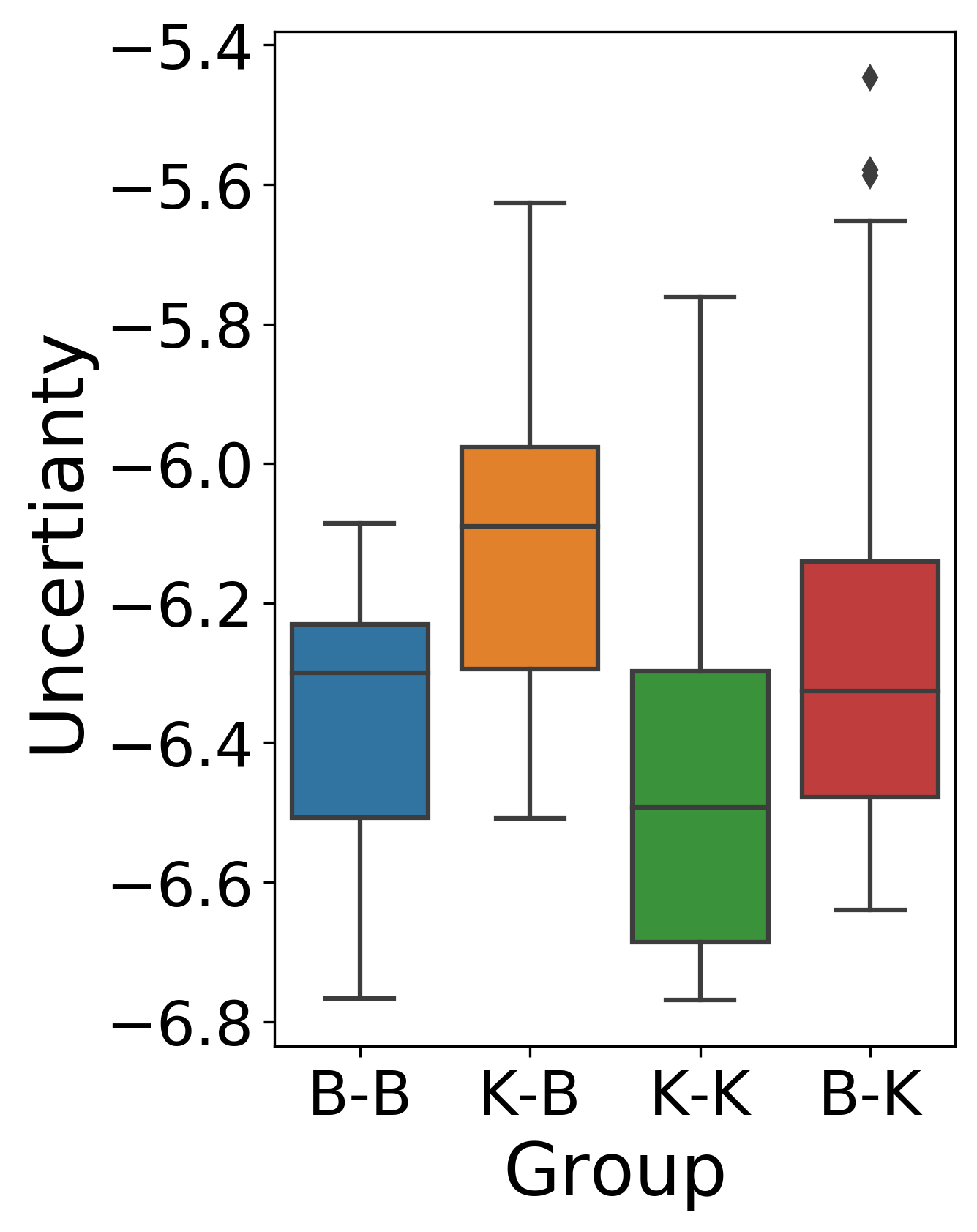}}
	\end{minipage}
	\begin{minipage}[b]{0.23\linewidth}
		\centering
		\subfloat[\label{fig5:b}\footnotesize Dropout]{
		\includegraphics[width=3cm]{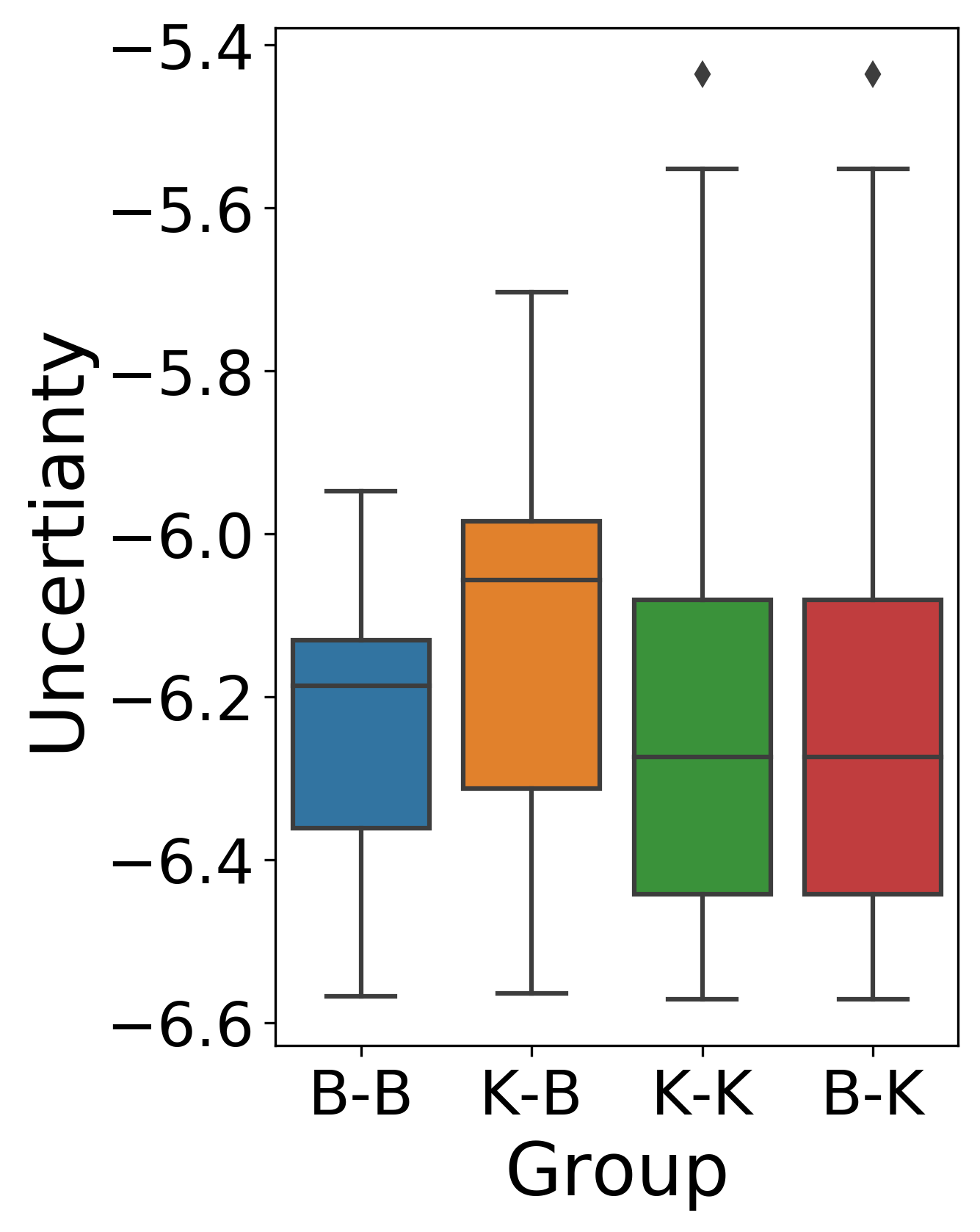}}
	\end{minipage}	
 	\begin{minipage}[b]{0.23\linewidth}
		\centering
		\subfloat[\label{fig5:c}\footnotesize SGLD]{
		\includegraphics[width=3cm]{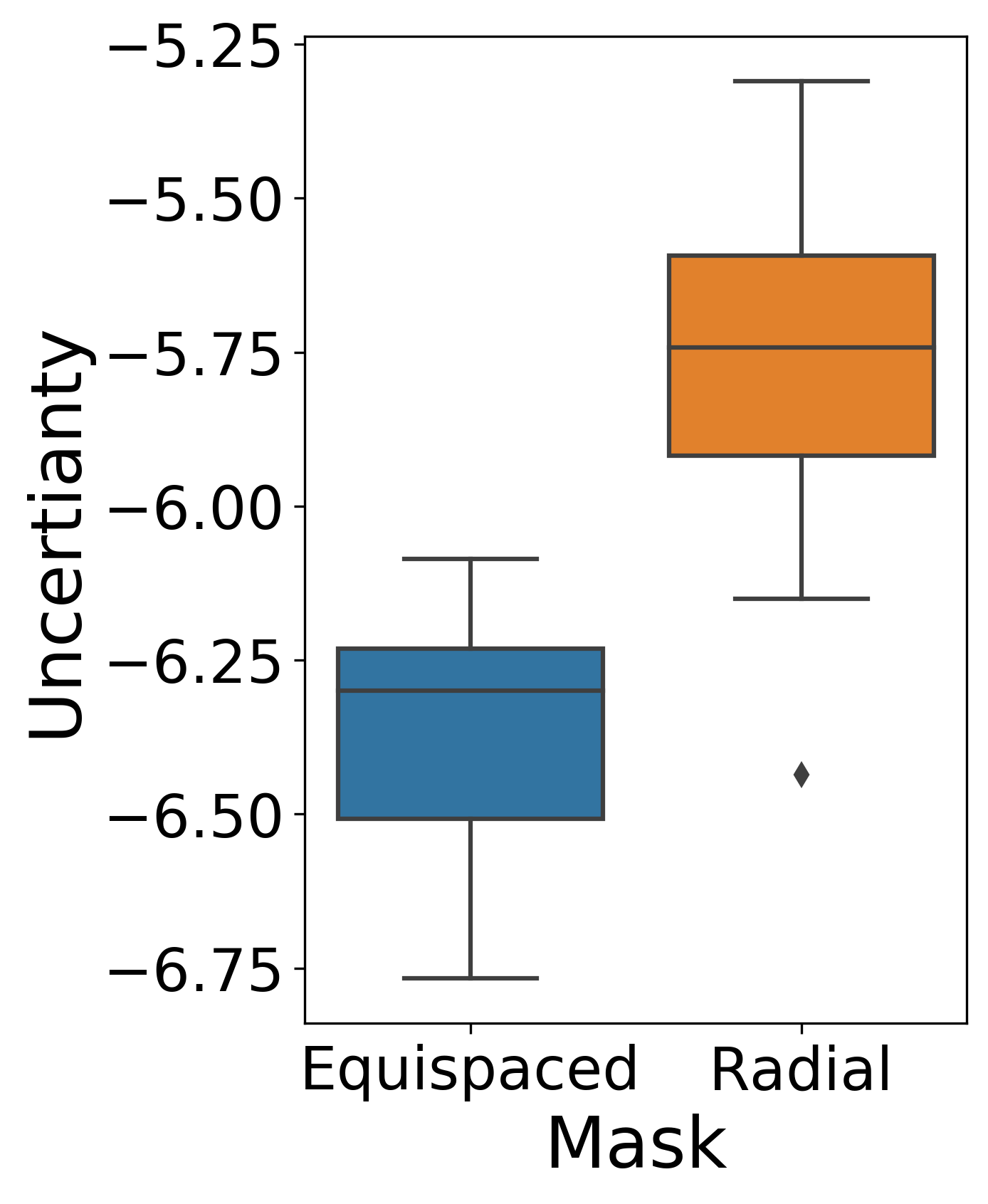}}
	\end{minipage}
	\begin{minipage}[b]{0.23\linewidth}
		\centering
		\subfloat[\label{fig5:d}\footnotesize Dropout]{
		\includegraphics[width=3cm]{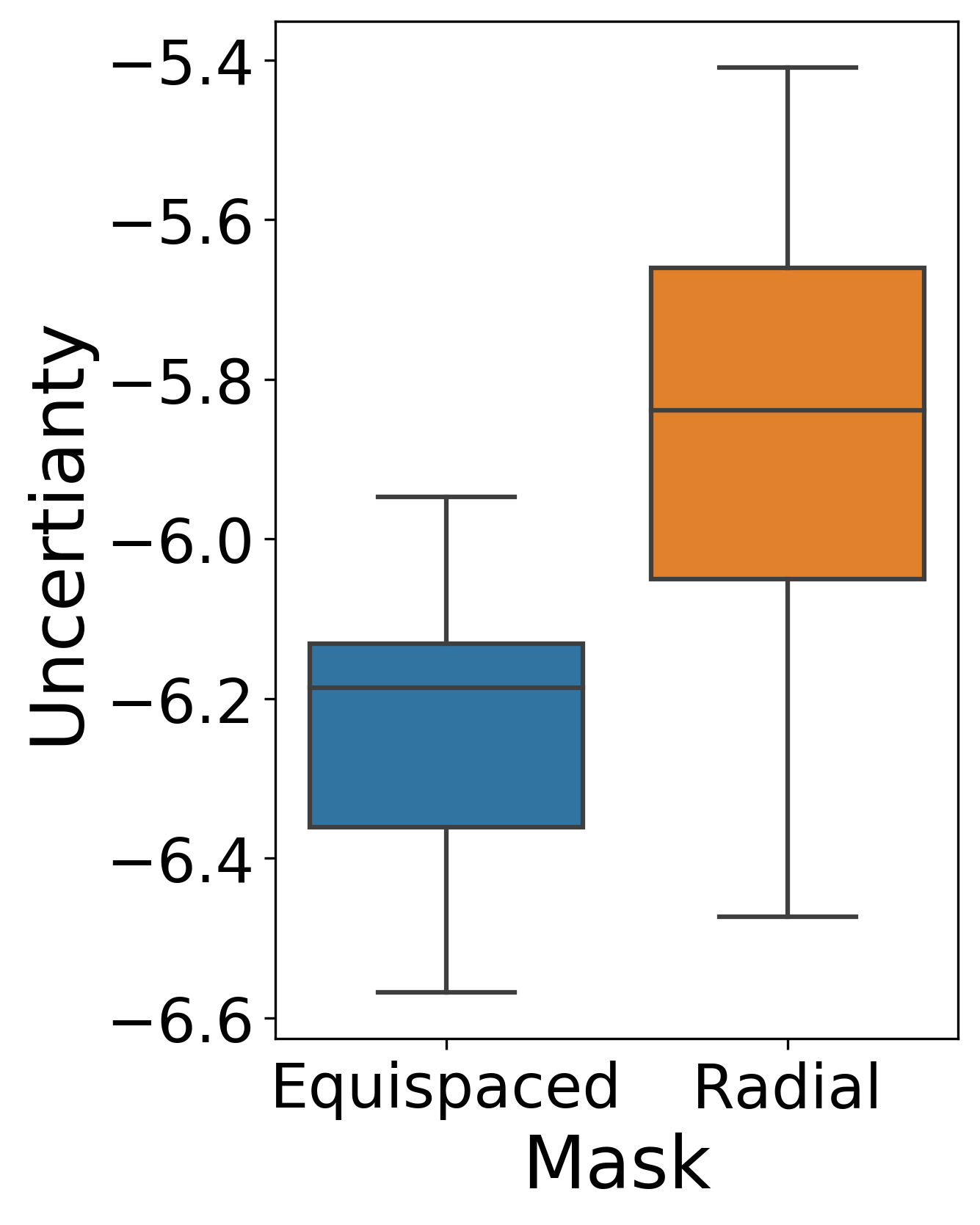}}
	\end{minipage}	

	\caption{Uncertainty Assessment. Box plots of the mean value of Std. estimate obtained by NPB-REC, in log scale, for different anatomical shifts \protect\subref{fig5:a} and \protect\subref{fig5:b} and two types of masks \protect\subref{fig5:c} and \protect\subref{fig5:d} for both NPB-REC and Dropout methods, respectively.  }\label{fig:uncBoxplots}
\end{figure}
\begin{figure*}[ht!]
\centering
	\begin{minipage}[b]{0.18\linewidth}
		\centering
		\includegraphics[width=2.3cm]{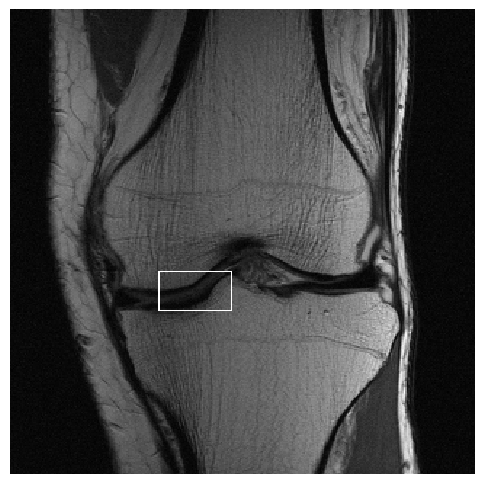}
	\end{minipage} 
	\begin{minipage}[b]{0.18\linewidth}
		\centering
		\includegraphics[width=2.3cm]{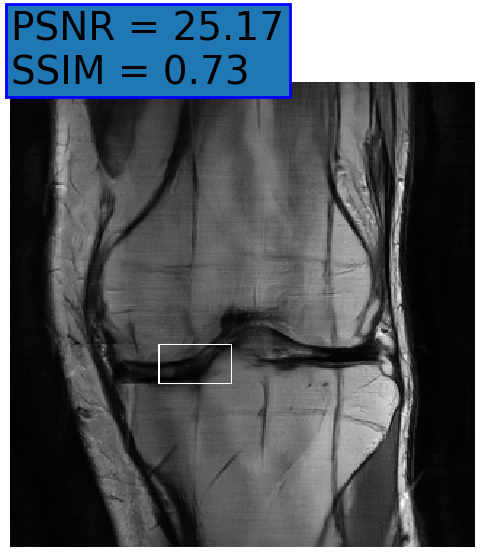}
	\end{minipage}
	\begin{minipage}[b]{0.18\linewidth}
		\centering
		\includegraphics[width=2.3cm]{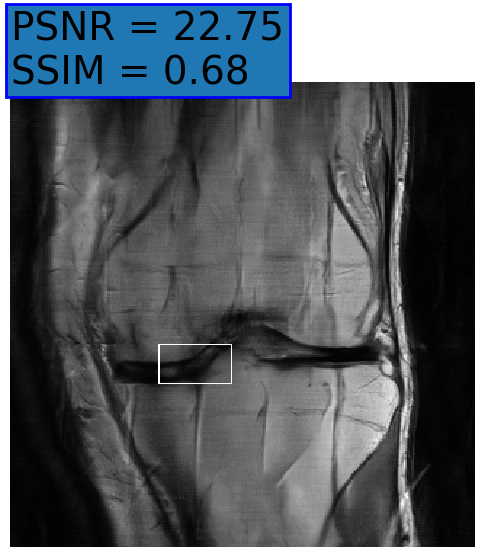}
	\end{minipage} 
	\begin{minipage}[b]{0.18\linewidth}
		\centering
		\includegraphics[width=2.3cm]{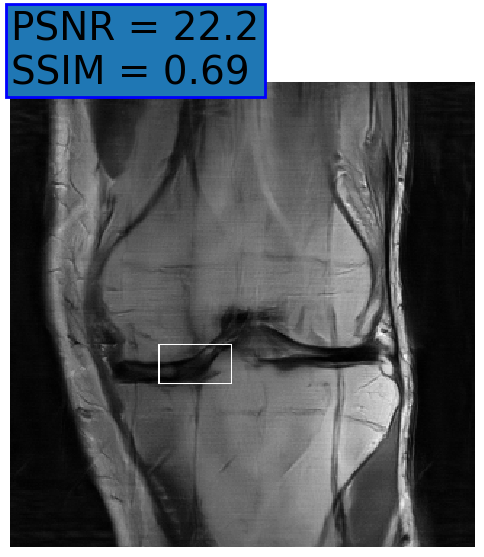}
	\end{minipage} 
		\begin{minipage}[b]{0.18\linewidth}
		\centering
		\includegraphics[width=2.3cm]{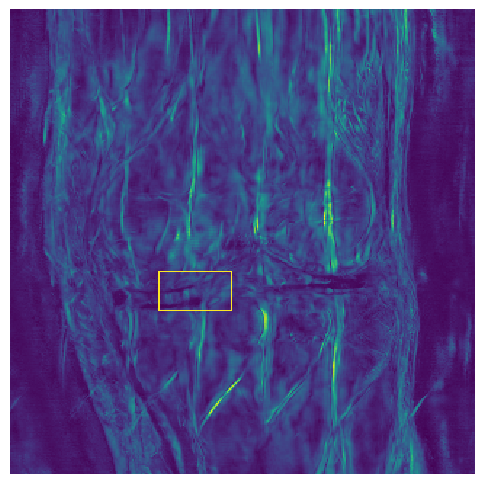}
	\end{minipage}
	\\
	\begin{minipage}[b]{0.18\linewidth}
		\centering
		\includegraphics[width=2.3cm]{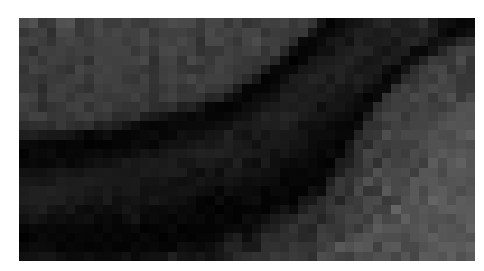}
	\end{minipage} 
	\begin{minipage}[b]{0.18\linewidth}
		\centering
		\includegraphics[width=2.3cm]{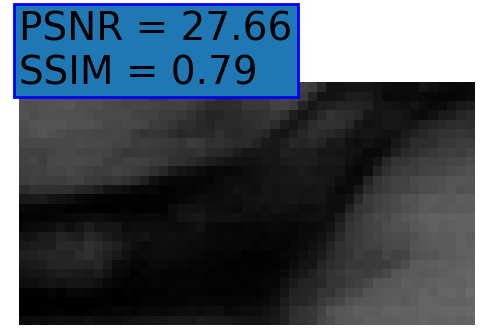}
	\end{minipage} 
		\begin{minipage}[b]{0.18\linewidth}
		\centering
		\includegraphics[width=2.3cm]{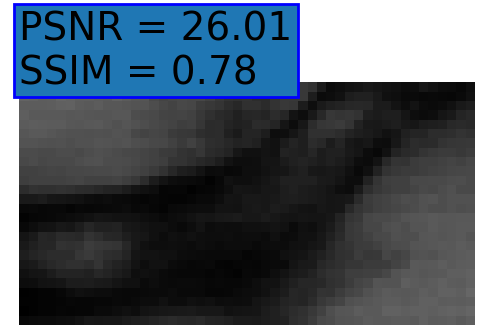}
	\end{minipage} 
		\begin{minipage}[b]{0.18\linewidth}
		\centering
		\includegraphics[width=2.3cm]{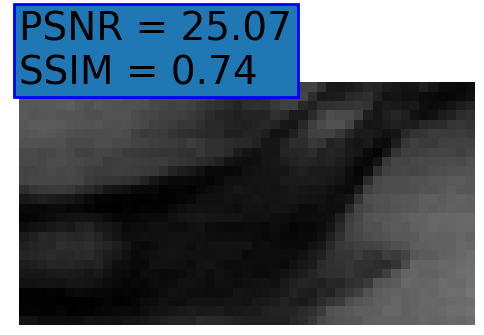}
	\end{minipage} 
		\begin{minipage}[b]{0.18\linewidth}
		\centering
		\includegraphics[width=2.3cm]{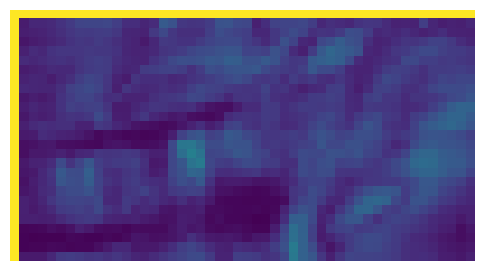}
	\end{minipage}
    \\
	\begin{minipage}[b]{0.18\linewidth}
		\centering
		\includegraphics[width=2.3cm]{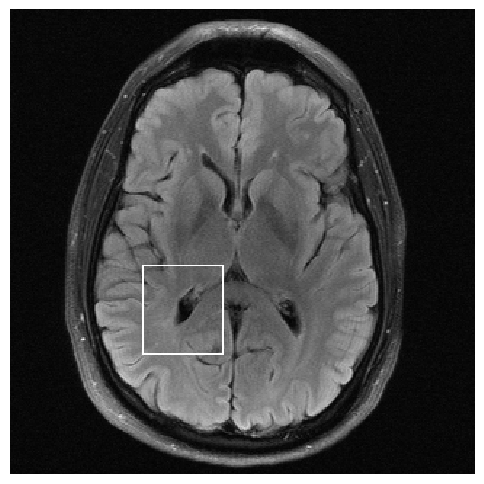}
	\end{minipage} 
	\begin{minipage}[b]{0.18\linewidth}
		\centering
		\includegraphics[width=2.3cm]{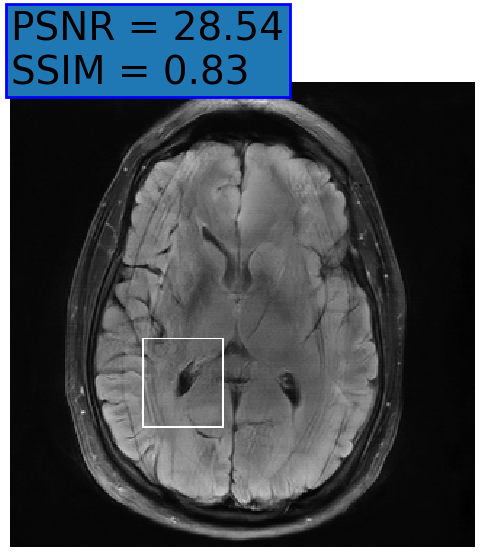}
	\end{minipage}
	\begin{minipage}[b]{0.18\linewidth}
		\centering
		\includegraphics[width=2.3cm]{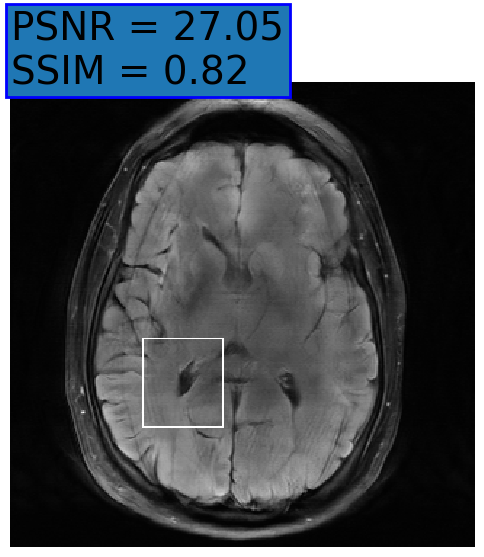}
	\end{minipage} 
	\begin{minipage}[b]{0.18\linewidth}
		\centering
		\includegraphics[width=2.3cm]{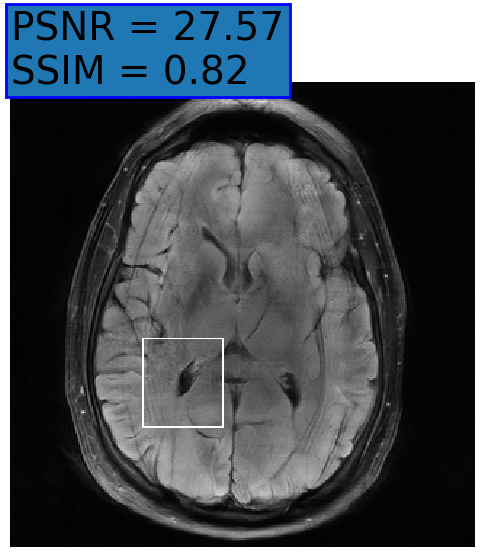}
	\end{minipage} 
		\begin{minipage}[b]{0.18\linewidth}
		\centering
		\includegraphics[width=2.3cm]{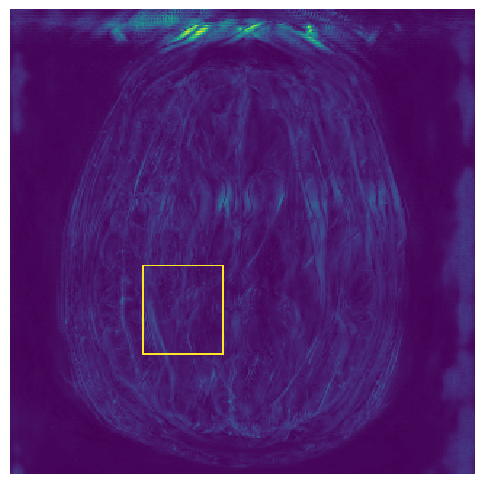}
	\end{minipage}
	\\
		\begin{minipage}[b]{0.18\linewidth}
		\centering
		\includegraphics[width=2.3cm]{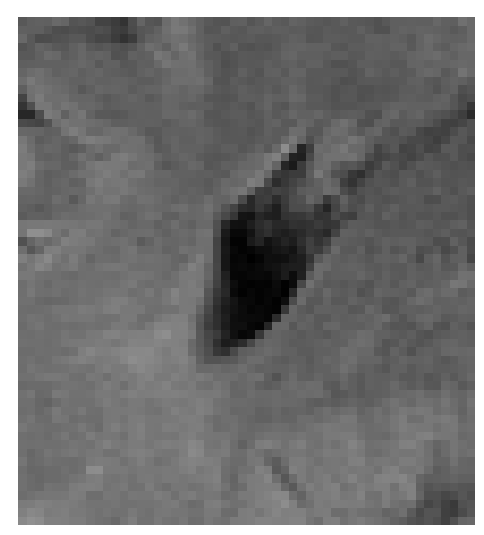}
	\end{minipage} 
	\begin{minipage}[b]{0.18\linewidth}
		\centering
    \includegraphics[width=2.3cm]{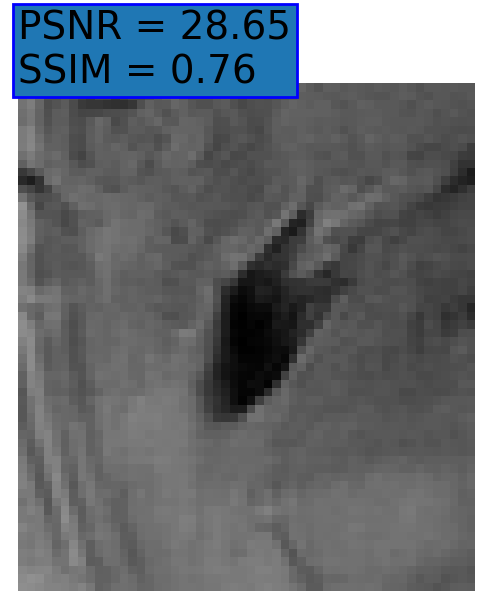}
	\end{minipage} 
		\begin{minipage}[b]{0.18\linewidth}
		\centering
	    \includegraphics[width=2.3cm]{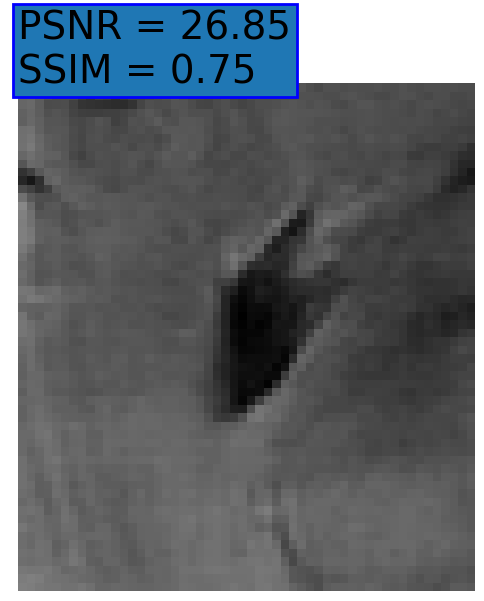}
	\end{minipage} 
		\begin{minipage}[b]{0.18\linewidth}
		\centering
	    \includegraphics[width=2.3cm]{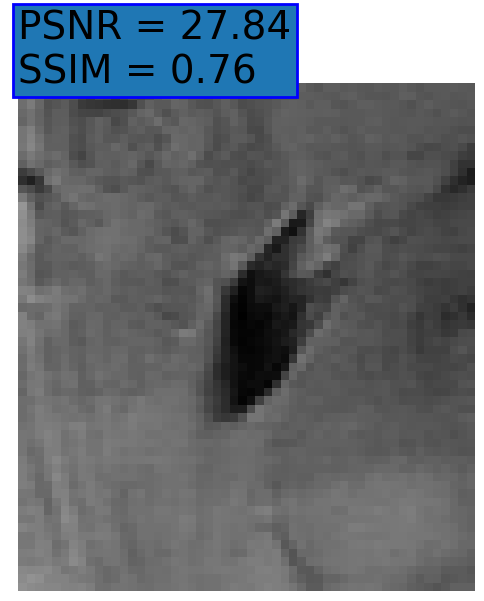}
	\end{minipage} 
		\begin{minipage}[b]{0.18\linewidth}
		\centering
		\includegraphics[width=2.3cm]{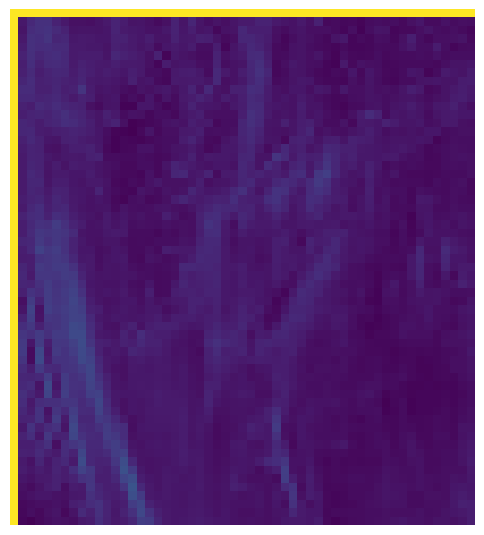}
	\end{minipage}
		\begin{minipage}[b]{0.18\linewidth}
		\centering
		\includegraphics[width=2.3cm]{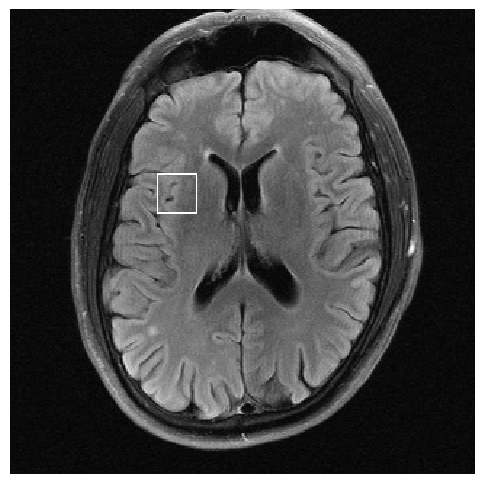}
	\end{minipage} 
	\begin{minipage}[b]{0.18\linewidth}
		\centering
		\includegraphics[width=2.3cm]{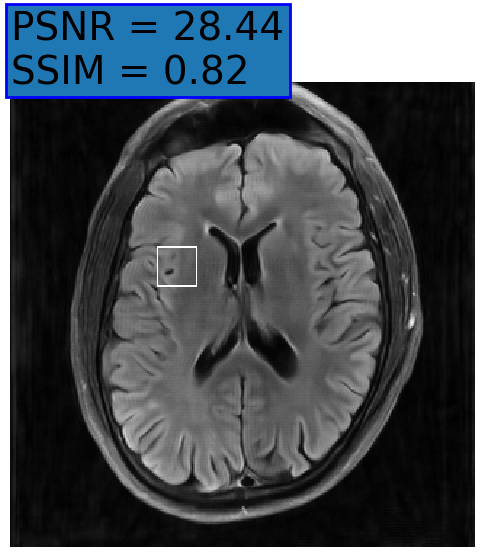}
	\end{minipage}
	\begin{minipage}[b]{0.18\linewidth}
		\centering
		\includegraphics[width=2.3cm]{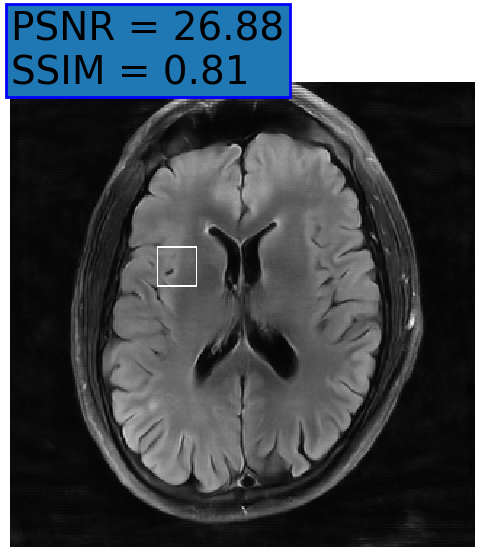}
	\end{minipage} 
	\begin{minipage}[b]{0.18\linewidth}
		\centering
		\includegraphics[width=2.3cm]{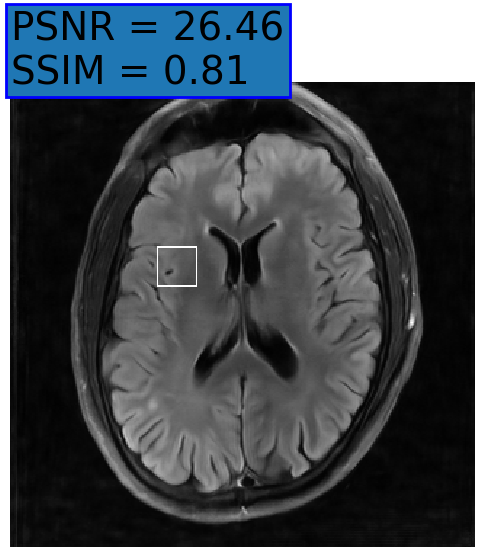}
	\end{minipage} 
		\begin{minipage}[b]{0.18\linewidth}
		\centering
		\includegraphics[width=2.3cm]{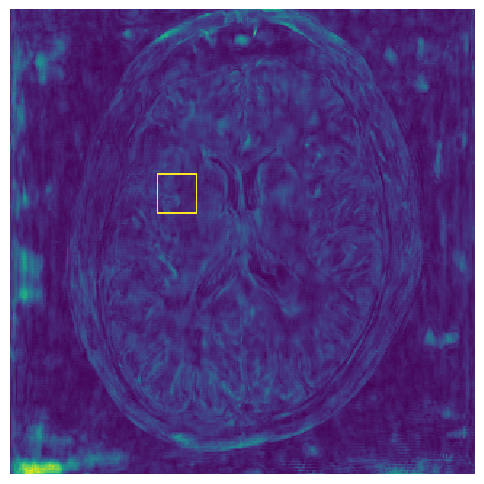}
	\end{minipage}
	\\
		\begin{minipage}[b]{0.18\linewidth}
		\centering
		\subfloat[\label{fig3:a}\footnotesize GT]{\includegraphics[width=2.3cm]{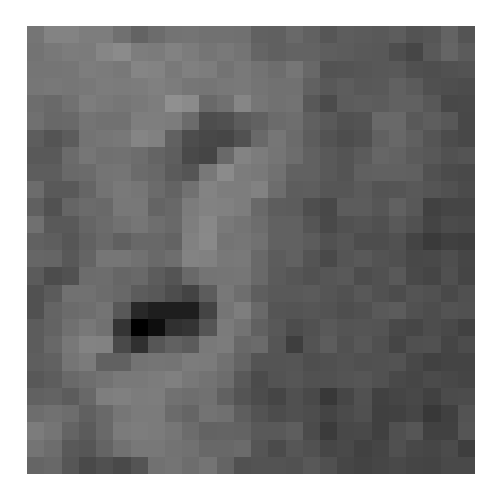}}
	\end{minipage} 
	\begin{minipage}[b]{0.18\linewidth}
		\centering
		\subfloat[\label{fig3:b}\footnotesize NPB-REC Avg.]{\includegraphics[width=2.3cm]{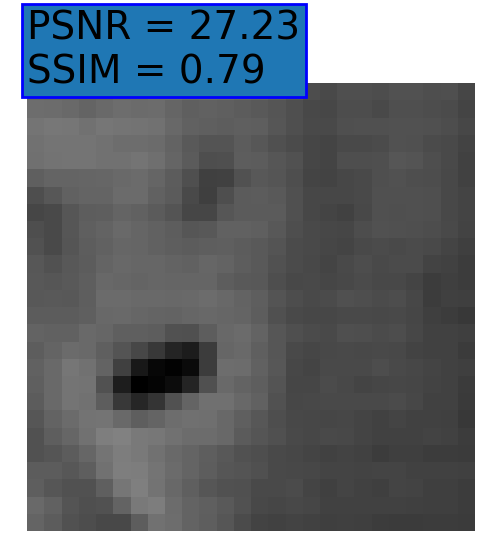}}
	\end{minipage} 
		\begin{minipage}[b]{0.18\linewidth}
		\centering
		\subfloat[\label{fig3:c}\footnotesize baseline]{\includegraphics[width=2.3cm]{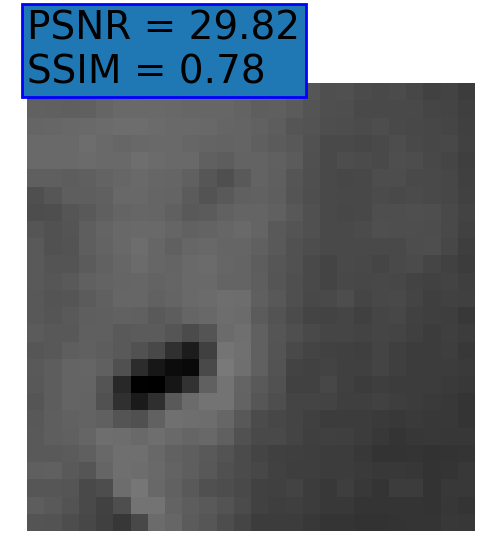}}
	\end{minipage} 
		\begin{minipage}[b]{0.18\linewidth}
		\centering
		\subfloat[\label{fig3:d}\footnotesize Dropout]{\includegraphics[width=2.3cm]{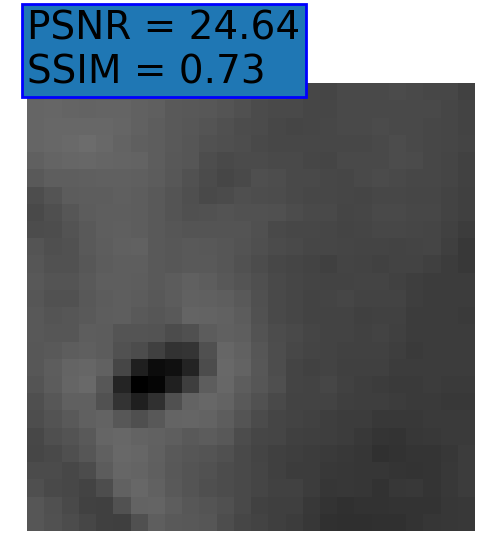}}
	\end{minipage} 
		\begin{minipage}[b]{0.18\linewidth}
		\centering
		\subfloat[\label{fig3:e}\footnotesize NPB-REC Std.]{\includegraphics[width=2.3cm]{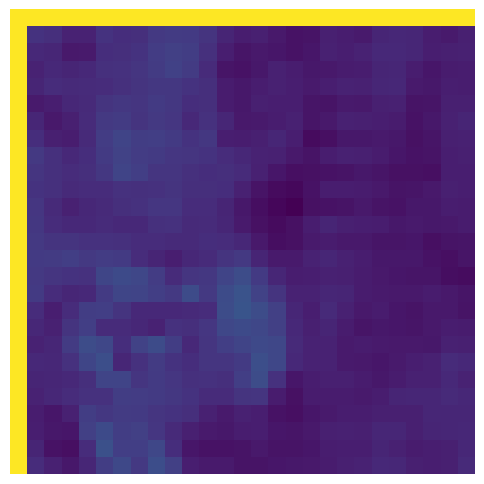}}
	\end{minipage}
	\caption{Examples of reconstruction results. Rows 1, 3, and 5: The reference (GT) fully sampled image, the reconstructed images obtained by the three models (1-3), NPB-REC, baseline, E2E-VarNet trained with Dropout, and the Std. map derived from our method for example from the knee dataset, the brain dataset, and the brain dataset sampled with radial mask $L=50$, respectively. Rows 2, 4, and 6: The corresponding annotated ROIs of the cartilage, edema, and nonspecific white matter lesion respectively. }\label{fig:RecGenMRIexample}	
\end{figure*} 

\section{Experimental Results}
\label{sec:expresults}
\subsection{Reconstruction accuracy}
Fig.~\ref{fig:RecMRIexample} presents examples of reconstruction results obtained by (1) our NPB-REC approach, (2) the baseline, and (3) Monte Carlo Dropout, for equispaced masks with two different acceleration rates $R=4$ and $R=8$.   The reconstructed images predicted by the three models are smoother than the reference image. This is due to the fact that all the models were trained with SSIM loss, which tends to produce overly smooth reconstructions while preserving the diagnostic content and the anatomical features \cite{sriram2020end}. These images can be enhanced by dithering the image by adding a small amount of random Gaussian noise to produce a more textured reconstruction, as proposed in \cite{sriram2020grappanet}.

 Table~\ref{tab:results} presents the mean PSNR and SSIM metrics, calculated over the whole inference set, for the three models. Our NPB-REC approach achieved significant improvements over the other methods in terms of PSNR and SSIM (Wilcoxon signed-rank test, p$\ll$1e-4, except for SSIM values in line W, $R=4$ where they are roughly the same for NPB-REC and Baseline). The improvement in the reconstruction performance can be noted both quantitatively from the metrics, especially for masks with acceleration rate $R=8$, and qualitatively via the images of annotations, where our results show less smoothness than those obtained by Dropout.
 \subsection{Uncertainty Assessment}
We calculated the mean value of the std. maps, obtained by our method and the Monte Carlo Dropout method, for all images in the inference set and utilized it as an uncertainty measure.
The correlation between these uncertainty measures and reconstruction error (MSE) is depicted in Fig.~\ref{fig:unc}. Our NPB-REC uncertainty measure exhibits a higher correlation with the Reconstruction error compared to Dropout (Pearson correlation coefficient of $r=0.93$ vs. $r=0.89$). Further, fig.~\ref{fig:unc}\protect\subref{fig3:c} demonstrates our uncertainty measures correlate better with the acceleration rates used during acquisition. 
These outcomes, in turn, indicate the ability of our uncertainty measure to detect unreliable reconstruction performance. It is worth mentioning that the strong linearity of the correlation exhibited was preserved even for higher acceleration rates and when we repeated the same experiments on the knee test set (in all cases we obtained $r>0.92$).     

Fig.~\ref{fig:uncBoxplots} presents the uncertainty values measured in the cases of anatomical and undersampling mask distribution shifts. For the anatomical distribution shift, we considered the following scenarios: brain test set when the predictive model is trained on brain images as well (B-B), the same model tested on knee data (K-B), training and testing the model on the knee dataset (K-K) and the same model tested on brain data (B-K). Our uncertainty measures were significantly higher for the out-of-distribution cases compared to the within-distribution cases for both anatomical and undersampling mask distributions. In contrast, the uncertainty measures of Dropout method don't show a significant difference in the case of anatomical shifts from knee to brain (see top-right part in Fig.~\ref{fig:uncBoxplots}). 

\begin{table*}[]
\caption{Reconstruction accuracy for the knee dataset. Rows top to bottom: PSNR and SSIM metrics calculated on the annotated anatomical ROIs (denoted by 'A'), and the whole physical images (denoted by 'W') with masks of acceleration rates $R=4$, $R=8$ and $R=12$, respectively.} \label{tab:kneeresults}
\setlength{\tabcolsep}{9pt}
\centering
\resizebox{\textwidth}{!}{%
\begin{tabular}{cccccccc}
\toprule
                    &   & \multicolumn{2}{c}{\textbf{NPB-REC}}          & \multicolumn{2}{c}{\textbf{Baseline}}     & \multicolumn{2}{c}{\textbf{Dropout}}      \\ \midrule
\textbf{R}                  &   & \textbf{PSNR}           & \textbf{SSIM }            & \textbf{PSNR }          & \textbf{SSIM}            & \textbf{PSNR}           & \textbf{SSIM}            \\ \midrule
\multirow{2}{*}{4}  & W & $36.21 \pm 5.57$ & $0.912 \pm 0.0963$ & $37.43 \pm 5.58$ & $0.912 \pm 0.098$ & $\mathbf{37.68 \pm 4.81}$ & $\mathbf{0.912 \pm 0.091}$ \\ \cmidrule(l){2-8}  
                    & A & $24.31 \pm 6.99$ & $0.855 \pm 0.214$  & $25.28 \pm 6.71$ & $0.861 \pm 0.215$ & $\mathbf{25.57 \pm 7.13}$ & $\mathbf{0.862 \pm 0.218}$ \\ \midrule
\multirow{2}{*}{8}  & W & $\mathbf{32.38 \pm 4.46}$ & $\mathbf{0.849 \pm 0.108}$  & $30.3 \pm 4.14$  & $0.825 \pm 0.103$ & $31.63 \pm 4.38$ & $0.836 \pm 0.096$ \\ \cmidrule(l){2-8}
                    & A & $\mathbf{20.77 \pm 6.18}$ & $\mathbf{0.716 \pm 0.332}$  & $19.02 \pm 5.86$ & $0.668 \pm 0.342$ & $20.18 \pm 5.74$ & $0.691 \pm 0.324$ \\  \midrule
\multirow{2}{*}{12} & W & $\mathbf{29.17 \pm 4.23}$ & $\mathbf{0.784 \pm0.104}$   & $25.3 \pm 5.66$  & $0.724 \pm 0.115$ & $28.23 \pm 4.89$ & $0.766 \pm 0.102$ \\ \cmidrule(l){2-8}
                    & A & $\mathbf{17.85 \pm 5.56}$ & $\mathbf{0.573 \pm 0.361}$  & $15.03 \pm 7.03$ & $0.448 \pm 0.414$ & $16.95 \pm 6.27$ & $0.534 \pm 0.364$
                    \\ \bottomrule
\end{tabular}
}
\end{table*}

\begin{table*}[]
\caption{Reconstruction accuracy for the brain dataset. Rows top to bottom: PSNR and SSIM metrics calculated on the annotated anatomical ROIs (denoted by 'A'), and the whole physical images (denoted by 'W') with masks of acceleration rates $R=4$, $R=8$ and $R=12$, respectively.} \label{tab:brainresults}
\setlength{\tabcolsep}{9pt}
\resizebox{\textwidth}{!}{%
\begin{tabular}{cccccccc}
                    &   & \multicolumn{2}{c}{\textbf{NPB-REC}}         & \multicolumn{2}{c}{\textbf{Baseline}}     & \multicolumn{2}{c}{\textbf{\textbf{Dropout}}}      \\ \midrule
R                   &   & \textbf{PSNR}           & \textbf{SSIM }           & \textbf{PSNR}           & \textbf{SSIM}            & \textbf{PSNR}           & \textbf{SSIM}            \\ \midrule
\multirow{2}{*}{4}  & W & $38.18 \pm 7.29$ & $0.928 \pm 0.091$ & $\mathbf{38.55 \pm 6.64}$ & $\mathbf{0.935 \pm 0.082}$ & $38.03 \pm 7.04$ & $0.928 \pm 0.090$ \\ \cmidrule(l){2-8}
                    & A & $\mathbf{29.72 \pm 8.68}$ & $0.870 \pm 0.211$ & $29.4 \pm 8.12$  & $\mathbf{0.871 \pm 0.202}$ & $29.45 \pm 8.62$ & $0.867 \pm 0.211$ \\ \midrule
\multirow{2}{*}{8}  & W & $\mathbf{31.93 \pm 5.56}$ & $\mathbf{0.875 \pm 0.105}$ & $30.93 \pm 5.07$ & $0.865 \pm 0.100$ & $31.33 \pm 5.20$ & $0.869 \pm 0.102$ \\\cmidrule(l){2-8}
                    & A & $\mathbf{22.43 \pm 7.47}$ & $\mathbf{0.668 \pm 0.363}$ & $21.74 \pm 6.72$ & $0.649 \pm 0.345$ & $22.03 \pm 7.06$ & $0.658 \pm 0.355$ \\ \midrule
\multirow{2}{*}{12} & W & $\mathbf{28.07 \pm 5.00}$ & $\mathbf{0.817 \pm 0.106}$ & $27.43 \pm 5.00$ & $0.796 \pm 0.107$ & $28.07 \pm 5.00$ & $0.817 \pm 0.106$ \\ \cmidrule(l){2-8}
                    & A & $\mathbf{19.75 \pm 6.40}$ & $\mathbf{0.541 \pm 0.336}$ & $19.25 \pm 6.36$ & $0.518 \pm 0.303$ & $19.56 \pm 6.31$ & $0.534 \pm 0.320$ \\ \bottomrule
\end{tabular}%
}
\end{table*}
\begin{table*}[]
\caption{Reconstruction accuracy for radial undersampling. Rows top to bottom: PSNR and SSIM metrics calculated on the annotated anatomical ROIs (denoted by 'A'), and the whole physical images (denoted by 'W') with \textit{radial} masks of number of lines $L=50$ and $L=40$, respectively.} \label{tab:radialresults}
\setlength{\tabcolsep}{9pt}
\resizebox{\textwidth}{!}{
\begin{tabular}{cccccccc}
                    &   & \multicolumn{2}{c}{\textbf{NPB-REC}}            & \multicolumn{2}{c}{\textbf{Baseline}}        & \multicolumn{2}{c}{\textbf{Dropout}}          \\ \\ \midrule
L                   &   & \textbf{PSNR }           & \textbf{SSIM}              & \textbf{PSNR}            & \textbf{SSIM}              & \textbf{PSNR}            & \textbf{SSIM}               \\\\ \midrule 
\multirow{2}{*}{50} & W & $\mathbf{34.21 \pm 5.07}$  & $\mathbf{0.8867 \pm 0.1026}$ & $33.86 \pm 5.69$  & $0.8827 \pm 0.1107$ & $33.67 \pm 5.049$ & $0.8839 \pm 0.09501$ \\  \cmidrule(l){2-8}  
                    & A & $25.95 \pm 6.042$ & $0.8064 \pm 0.2259$ & $\mathbf{27.23 \pm 6.194}$ & $\mathbf{0.8139 \pm 0.2222}$ & $26.95 \pm 6.068$ & $0.8023 \pm 0.2246$  \\ \midrule
\multirow{2}{*}{40} & W & $\mathbf{33.28 \pm 4.942}$ & $\mathbf{0.8745 \pm 0.1045}$ & $33.14 \pm 5.61$  & $0.8727 \pm 0.1154$ & $32.81 \pm 4.949$ & $0.8704 \pm 0.09966$ \\ \cmidrule(l){2-8}
                    & A & $25.01 \pm 6.362$ & $0.7775 \pm 0.2731$ & $\mathbf{26.17 \pm 6.568}$ & $\mathbf{0.7796 \pm 0.2715}$ & $25.77 \pm 6.213$ & $0.7649 \pm 0.2764$ \\ \bottomrule
\end{tabular}
}
\end{table*}

\subsection{Improved Generalization Capability}
Fig.~\ref{fig:RecGenMRIexample} depicts the reconstruction results obtained by (1) our NPB-REC approach, (2) the baseline, and (3) Monte Carlo Dropout, for examples of both the knee (row 1) and the brain (rows 3 and 5) datasets. The first two examples were obtained by the models that trained on brain data and were tested on knee data (row 1) and vice versa (row 3). In knee images, all three models exhibit minimal differences in handling pronounced folding artifacts. However, our approach demonstrates superior reconstruction performance for anatomical annotations or pathological regions, labeled by radiologists, in terms of both PSNR and SSIM.
The last example is a reconstructed brain image that was obtained by the three models, where the k-space input is undersampled by a radial mask with $L=50$. 
Table~\ref{tab:kneeresults} presents the mean PSNR and SSIM metrics, calculated over the whole inference knee dataset, for the three models that were trained on brain data. The NPB-REC method shows a considerable improvement in the generalization ability on knee data for higher acceleration rates $R=8$ and $R=12$ in both annotation regions and the whole images. However, both the baseline and Dropout methods show a slight improvement in the metrics in the case of a small acceleration rate $R=4$. 
Table~\ref{tab:brainresults} shows the results of the inverse experiment, i.e. calculating the mean PSNR and SSIM metrics over the whole brain inference set for the three models that were trained on knee data only. The generalization capability of the three models in the case of an anatomical distribution shift from the knee to the brain is better than the opposite scenario. This is not only measured quantitatively by the SSIM and PSNR metrics, but also visible in the predicted reconstructions. For instance, the brain MRI reconstructed images exhibit fewer artifacts than the produced knee images, as shown in row 1 vs. row 3 in Fig.~\ref{fig:RecGenMRIexample}. 

Table~\ref{tab:brainresults} presents the mean PSNR and SSIM metrics calculated over the whole inference knee dataset that were obtained by the three models, where the k-space input is undersampled by a radial mask with $L=50$ and $L=40$. Although the NPB-REC approach improves the accuracy of reconstruction when the metrics are measured on the whole image, it doesn't yield the best performance on the annotation ROIs. However, it is still able to reconstruct high-quality images while preserving the important anatomical content in these annotations. This can be clearly observed in the last row in Fig.~\ref{fig:RecGenMRIexample}, where the topmost part of the ROI image predicted by NPB-REC is preserved, in contrast to the other benchmarks.     

\section{Discussion}
 DNN-based models have been successfully employed to reconstruct high-resolution MRI images from undersampled k-space measurements \cite{hammernik2018learning,shaul2020subsampled,eo2018kiki,akccakaya2019scan,putzky2019rim,quan2018compressed,radford2015unsupervised}. However, their practical use in clinical settings remains questionable due to the lack of uncertainty assessment. Uncertainty quantification is crucial in DNN-based image reconstruction models to facilitate reliable clinical decision-making based on the reconstructed images.

Bayesian models enable safer utilization of DNN methods in neuroimaging studies, improve generalization, and enable the quantification of uncertainty \cite{edupuganti2020uncertainty}. In this work, we proposed "NPB-REC," a non-parametric fully Bayesian framework for uncertainty assessment in MRI reconstruction from undersampled "k-space" data. To achieve this, we employed the Stochastic Gradient Langevin Dynamics (SGLD) \cite{cheng2019bayesian,welling2011bayesian} algorithm during the training of our model to characterize the posterior distribution of the network parameters. This, in turn, enables sampling from the posterior and estimation of the uncertainty in MRI reconstruction. Additionally, this strategy improves the generalization capability of the reconstruction system due to the incorporation of SGLD in the training process \cite{neelakantan2015adding}.

Our method is not limited to specific DNN architectures, in contrast to other Bayesian methods such as VAEs and Monte Carlo dropout \cite{tezcan2018mr,edupuganti2020uncertainty,cai2019multi}. Further, it is a non-parametric strategy for distribution characterization of the predictions, in the sense that it doesn't assume a parametric distribution like the form of a Gaussian distribution, which may represent an oversimplification of the unknown true underlying distribution. However, it does require additional minor storage space to save the network parameters obtained in the last iterations of the training. 

Our experimental results demonstrated that the incorporation of our Bayesian method has significantly boosted the performance in terms of reconstruction accuracy compared to the baseline, even for higher acceleration rates. Overall, the proposed NPB-REC method improves generalization capabilities, both in scenarios with high acceleration rates and anatomical shifts. Additionally, it allows the assessment of uncertainty in the MRI reconstruction and provides a principled mechanism for determining out-of-distribution data. In contrast to Dropout method, our uncertainty measures show a significant difference in both cases of anatomical shifts, from knee to brain and vice versa. This implies that the uncertainty measure derived from the Dropout method may not be effective in identifying anatomical shifts.  
These outcomes demonstrate the ability of our uncertainty measures to detect out-of-distribution cases. Furthermore, our method demonstrates superior generalization on knee data compared to Dropout, as evidenced by higher PSNR ($31.93$ vs. $31.33$) and SSIM ($0.87$ vs. $0.86$) for an acceleration rate of $R=8$, highlighting its enhanced robustness.
    
The difference in performance, as measured by SSIM and PSNR metrics, between the anatomical distribution shift from knee to brain MRI reconstruction and the reverse scenario is primarily attributed to the model's ability to generalize more effectively when trained on knee data and applied to brain data. This experimental observation, as we deposit,  stems from the inherent distinctions in characteristics between brain and knee MRI images, including anatomical structures, motion artifacts, and imaging parameters.
For instance, brain MRI images typically exhibit a more uniform and smoother appearance, while knee MRI images often display a greater complexity with variable features. Consequently, the model may encounter challenges when adapting to these inherent dissimilarities.

Lastly, in terms of computational complexity, our NPB-REC, the baseline, and Monte-Carlo Dropout methods have the same training time and runtime for reconstruction prediction. However, to enable uncertainty quantification by sampling from the posterior, one needs to perform $9$ forward passes to the reconstruction system. This additional sampling process increases the overall runtime by a factor of $9$. This applies to both our NPB-REC and Monte-Carlo Dropout methods.

\section{Conclusions}
We presented NPB-REC, a non-parametric Bayesian approach for reconstructing MRI images from under-sampled k-space data with uncertainty estimation. Our method utilizes noise injection for efficient sampling of the true posterior distribution of the network parameters during training, and can be incorporated into any existing network architecture. Our experiments demonstrate that our approach offers improved uncertainty quantification that is more strongly correlated with reconstruction error compared to the Monte Carlo Dropout method. Additionally, our method exhibits greater robustness to various distribution shifts, such as changes in the anatomical region or undersampling mask, and outperforms state-of-the-art techniques in terms of reconstruction quality, particularly at acceleration rates higher than those used in training. These results suggest that NPB-REC has the potential to facilitate a safer adoption of DNN-based methods for MRI reconstruction from under-sampled data in clinical settings.



\section*{Acknowledgment}
Khawaled, S. is a fellow of the Ariane de Rothschild Women Doctoral Program. 
\section*{Statements and Declarations}
\subsection*{Conflict of Interest}
The authors have no conflicts of interest to declare.
\subsection*{Statement of Ethics}
No ethics approval was required since this research does not involve any human or animal subjects.
\subsection*{Funding Sources}
None
\subsection*{Data Availability Statement}
The data required to reproduce these findings is publicly available as referenced in the manuscript. The code and models to reproduce the findings is available at: \url{https://github.com/samahkh/NPB-REC}.
\subsection*{Author Contributions}
S.K. and M.F. contributed to the study's conception and design. S.K. developed the models and conducted the experiments. S.K. and M.F. analyzed the data. S.K. wrote the first draft of the manuscript, and then M.F. revised it. All authors read and approved the final manuscript.

\bibliographystyle{model2-names.bst}
\biboptions{authoryear}
\bibliography{refs.bib}

\end{document}